\documentclass[11pt]{article}

\usepackage{amssymb,amsfonts,amsmath,amsthm,amscd,dsfont,mathrsfs}
\usepackage{graphicx,float,psfrag,epsfig}
\usepackage{wrapfig}
\usepackage{hyperref}
\usepackage{algorithm,algorithmic}

\usepackage{color}

\DeclareMathAlphabet{\mathpzc}{OT1}{pzc}{m}{it}

\newtheorem{propo}{Proposition}[]
\newtheorem{lemma}{Lemma}
\newtheorem{definition}[propo]{Definition}

\newtheorem{thm}[propo]{Theorem}

\usepackage{pgfplots}
\usepackage{tikz}
\usepackage{subcaption}

\date{}

\def\L{\widetilde{L}}

\def\cH{\mathcal{H}}
\def\bH{\mathbb{H}}
\def\cT{\mathcal{T}}
\def\bT{\mathbb{T}}

\def\i{i^\prime}

\def\dmax{{\rm d_{max}}}

\def\Tr{{ \rm Tr}}
\def\diag{{\rm diag}}
\def\E{{\mathbb{E}}}

\def\reals{{\mathbb{R}}}

\def\<{\langle}
\def\>{\rangle}

\def\ones{{\mathds 1} }

\def\hTheta{{\widehat \Theta}}

\def\bb{{b}}
\def\hb{\widehat{b}}
\def\tb{\widetilde{b}}
\def\aa{{a}}

\def\mmm{{m}}

\newcommand{\norm}[1]{\|#1\|}

\def\wgt{{\omega}}

\def\diag{{\rm diag}}

\def\cc{{\tt cc}}
\def\hcc{{\widehat{\cc}}}

\begin{document}
\title{Number of Connected Components in a Graph:\\
	Estimation via Counting Patterns}

\author{
Ashish Khetan, Harshay Shah, and  Sewoong Oh\thanks{Author emails are: ashish.khetan09@gmail.com,hrshah4@illinois.edu,swoh@illinois.edu}\\
UIUC\\
}

\maketitle

\begin{abstract}
	Due to the limited resources and the scale of the graphs in modern datasets,
	we often get to observe a sampled subgraph of
	a larger original graph of interest, whether it is the worldwide web that has been crawled or
	social connections that have been surveyed.
	Inferring a global property of the original graph from such a
	sampled subgraph is of a fundamental interest. In this work, we focus on estimating the
	number of connected components.
	It is a challenging problem and, 
	for general graphs,  little is known about 
	the connection between 
	the observed subgraph 
	and the number of connected components of the original graph. 
	In order to make this connection, 
	we propose a highly redundant and large-dimensional representation of the subgraph,
	which at first glance seems counter-intuitive.
	A subgraph is represented by the counts of patterns, known as network motifs.
	This representation is crucial in  introducing a  novel   estimator for the number of connected components for
	general graphs, under the knowledge of the spectral gap of the original graph. 
	The connection is made precise via the Schatten $k$-norms of the graph Laplacian
	and the spectral representation of the number of connected components.
	We provide a guarantee on the resulting mean squared error that
	characterizes the bias variance tradeoff.
	Experiments on synthetic and real-world graphs suggest that we improve upon
	competing algorithms for graphs with spectral gaps bounded away from zero.  
\end{abstract}

%
%

\section{Introduction}
\label{sec:intro}

With the increasing size of modern datasets, a common network analysis task involves
 sampling a  graph,
due to restrictions on memory, communication, and computation resources.
From such a subgraph with sampled nodes and their interconnections,
we want to infer some global properties of the original graph that are relevant to the application in hand.
 This paper focuses on the task of inferring the number of connected components.
It is a fundamental graph property of interest in various applications such as
 estimating the weight of the minimum spanning trees \cite{CRT05,BKM14},
  estimating the number of classes in a population \cite{Goo49}, and visualizing large networks \cite{Raf05}.

In the sampled subgraph, the count
 of connected components
in general can be  smaller  as well as larger than that of the original graph.
Some connected components might not be sampled at all,
whereas  the connected nodes in the original graph is not guaranteed to be connected in the subgraph.
It is not clear
how the true number of components is related to the complex structure of the sampled graph.
For general graphs, it is unknown how to unravel the complex relationship between
the sampled subgraph and the global property of interest. 
In this paper, we propose encoding
the sampled subgraph by counting patterns in
the subgraph, and show that it
makes its connection to the
number of connected components  transparent.

We represent a graph by a vector of counts of all possible patterns, also known as network motifs.
For example,
the first and second entries  in this count vector encodes the number of nodes
and (twice) the number of edges, respectively.
Later entries encode the count of increasingly complex patterns:
the number of times a pattern is repeated in the graph.
This vector is clearly a redundant  over-representation whose dimension scales
super exponentially in the graph size. 
Perhaps surprisingly, for the purpose of approximately inferring a global property,
it suffices to have the first few hundred dimensions of this vector, corresponding to the counts of small patterns.
For counting those  patterns, we introduce novel  algorithms,
and give a precise characterization of
how the complexity (the size of the patterns included in the estimation)
trades off with accuracy (the mean squared error).

%


\bigskip
\noindent
{\bf Problem statement and our proposed approach.}
We want to estimate the number of connected components in a simple graph $G = (V,E)$ from  a sampled subset of its nodes and
the corresponding subgraph.
Let $N$ be the number of vertices and $\cc(G)$ the number of connected components in $G$.
We consider the subgraph sampling model, that is, a subset of vertices is sampled at random and the induced subgraph is observed.
We consider a Bernoulli sampling model, where each vertex is sampled independently with a probability $p$. Let $\Omega$ be the set of randomly observed vertices, and $G_{\Omega}$ be the corresponding induced subgraph,
i.e.~$G_\Omega=(\Omega,E_\Omega)$ where $(i,j)\in E_\Omega$ if $i,j \in\Omega $ and $(i,j)\in E$.
We want to estimate $\cc(G)$ from $G_{\Omega}$.
We propose a novel spectral approach, which
makes transparent the relation between the counts of patterns
and the number of connected components.

We propose characterizing  the number of connected components as
 the  count of zero eigenvalues of its Laplacian matrix $L \in \reals^{n \times n}$ given by
\begin{eqnarray} \label{eq:eq1}
	L  \;\; \equiv \;\; D - A \;,
\end{eqnarray}
where $D=\diag(A\ones)$ is the diagonal matrix of the degrees, and $A$ is the adjacency matrix of the graph $G$. The
rank of $L$ reveals $\cc(G)$  as
\begin{eqnarray} \label{eq:eq2}
\cc(G) &=& N - {\rm rank}(L) \nonumber\\
	&=& N - \sum_{i \in[N]} \mathbb{I}\big[\sigma_i(L) > 0 \big] \;,
\end{eqnarray}
where the $\sigma_i(L)$'s are
the singular values of the graph Laplacian $L$.
Using this relation directly for estimation is an overkill as estimating the singular values is more challenging than
estimating $\cc(G)$. Instead, we use a few steps of
functional approximations to relate to the pattern counts.
\textcolor{black}{By Gershgorin's circle theorem, we have $\sigma_i(L)\leq 2 \dmax$,
where $\dmax$ is the maximum degree in $G$.
We therefore normalize $L$ by $1/\beta $ for some $\beta\geq 2d_{\rm max}$ to ensure all eigenvalues lie in the unit interval $[0,1]$ and denote it by
$\L=(1/\beta)L$.}
For any constant $0 < \alpha < 1$
that separates the zero and non-zero eigenvalues such that
$\alpha < \min_{i}\{\sigma_{i}(\L): \sigma_i(\L) \neq 0 \}$,
we consider the following approximation of the rank function.
We approximate the step function in \eqref{eq:eq2} by a
continuous piecewise linear function $H_{\alpha} : [0,1] \rightarrow [0,1]$ illustrated in Figure \ref{fig:function}:
\begin{eqnarray} \label{eq:eq3}
	H_{\alpha}(x)& =& \begin{cases}
	1 & \text{if } x\in[\alpha,1] \;, \\
	\frac{x}{\alpha} & \text{if }x\in [0,\alpha]\;.
	\end{cases}
	\text{, \;\; and } \\
	\cc(G)  &=&  N - \sum_{i \in[N]} H_{\alpha} \big( \sigma_i(\L) \big)\;,	\nonumber
\end{eqnarray}
where we used the fact that the approximation is exact
under our assumption that the spectral gap is lower bounded by $\alpha$.
To connect it to the pattern counts, we propose a further approximation
using a polynomial function $f_{\alpha} : \reals \rightarrow \reals $ of a finite degree $m$.
Precisely, for $f_{\alpha}(x) =  a_1x + \cdots + a_m x^{m}$ (e.g.~Figure \ref{fig:function}),
we immediately have the following relation:
\begin{eqnarray} \label{eq:eq5}
	\sum_{i = 1}^N f_{\alpha}(\sigma_i(\L)) & = &  \sum_{k = 1}^m \frac{a_k}{\beta^k}\norm{L}_k^k\,,
	\label{eq:genrankapproximation}
\end{eqnarray}
where $\norm{L}_k^k$ is the Schatten-$k$ norm of $L$ which is defined as sum of $k$-th power of its singular values:
$\norm{L}_k^k \equiv  \sum_{i=1}^N \sigma_i(L)^k$.
As we choose $f_{\alpha} (x)$ to
be a close approximation of the desired $H_{\alpha}(x)$,
we have the following approximate relation:
$\cc(G)  \approx  N  - \sum_{k = 1}^m (a_k/\beta^k) \norm{ L}_k^k$,
which can be made arbitrarily close  by choosing a larger  degree $m$.

Finally, we propose using  the fact that
$\|L\|_k^k=\Tr(L^k)$ is a sum of the weights of all length $k$ closed walks.
Once we compute the (weighted) count of those walks for each pattern,
this gives a direct formula to approximate the number of connected components
from the counts. This approximation can be made as accurate as we want, by choosing the right order
$m$ in the polynomial approximation.
Unlike the singular values, the (weighted) counts can be directly
estimated from the sampled subgraph in a statistically efficient manner.
We introduce a novel unbiased estimator $\hTheta_k(G_\Omega)$ for  Schatten-$k$ norms of $L$   in Section \ref{sec:schatten}
that uses the  counts of patterns in the sampled subgraph,
and appropriately aggregates the estimated counts of the original graph.
Together with a  polynomial approximation $f_\alpha(x)$,
this gives a novel estimator:
\begin{eqnarray}
	\label{eq:eq7}
	\hcc(G_\Omega,\alpha,\beta,m) & \equiv & N   - \sum_{k=1}^m \frac{a_k}{\beta^k} \hTheta_k(G_\Omega)\,,
\end{eqnarray}
where $\hTheta_k(G_\Omega)$ is
an unbiased estimate of Schatten-$k$ norm of $L$
defined in \eqref{eq:estimate} and $a_k$'s are the coefficients in
the polynomial approximation
$f_\alpha(x) = a_1x + \cdots + a_m x^{m}$
as defined as in \eqref{eq:defpoly}.

\bigskip
\noindent
{\bf Related work.}
The connection between 
the number of connected components in the original graph 
and the counts of various patterns in the sampled graph has been explored in  
\cite{Fra78,Fra82} for limited classes of graphs with particular structures. 
These estimators are customized for
 two simple extreme cases of {\em forests} and {\em unions of disjoint cliques}, and 
rely only on  the counts of a few extremely simple patterns.

For a {\em forest} $G=(V,E)$,
the estimator introduced in   \cite{Fra78} exploits the simple relation that
the number of connected components is  $\cc(G) =|V|-|E|$.
Hence, we only need to estimate the number of edges.
This is a straightforward procedure that uses the counts of
{\em $k$-stars} in the sampled subgraph for $k\in\{0,1,\ldots \} $.
A $k$-star is a graph with one central node with $k$ adjacent nodes, mutually disjoint.

 For a {\em union of disjoint cliques} $G=(V,E)$,
 the estimator introduced in   \cite{Fra78} exploits the simple relation that
 the number of connected components is
 $\cc(G) = \sum_{k=1}^{|V|} \{\text{ \# of cliques of size $k$ }\}$.
 We only need to estimate the  number of cliques of each size $k$ in the original graph.
This is straightforward as
 the observed size of the cliques follow a multinomial distribution.
This  requires only  the counts of
{\em $k$-cliques} in the sampled subgraph for $k\in\{1,2,\ldots\}$ \cite{Goo49,Fra78}.
A $k$-clique is a  fully connected graph with $k$ nodes.
These approaches have recently been extended in \cite{KW17} to include  chordal graphs,
which introduces a novel idea of smoothing to achieve a strong performance guarantees.
However, none of these methods can be applied to our setting where
we consider the original graph to be a general graph.


\bigskip
\noindent
{\bf Contributions.}
We pose the problem of estimating the number of connected components as
a spectral estimation problem of estimating the rank of the graph Laplacian.
This is further split into two tasks of first estimating the Schatten $k$-norms
of the  Laplacian and then applying a functional approximation.

We propose an unbiased estimator of the Schatten $k$-norm $\|L\|_k^k$
based on the counts of patterns in the subsampled graph, known as
$k$-cyclic pseudographs.
The main challenge is in estimating the diagonal entries of $L$
(which is the degree of each node), that is critical in computing
the weighted counts of the $k$-cyclic pseudographs.
To overcome this challenge, in Section~\ref{sec:schatten_theta} 
we introduce an estimator that uses a novel idea of partitioning
the subsampled graph and stitching the estimated degrees in
each partition together.

Combining the estimated Schatten norms with polynomial approximation of $H_\alpha(x)$ in \eqref{eq:eq3},
we introduce a novel estimator of the number of connected components.
To the best of our knowledge, this is the first estimator 
with theoretical guarantees for general graphs. 
We provide a sharp characterization of the bias-variance tradeoff of our estimator in Section \ref{sec:main}.
Numerical experiments on synthetic and real-world graphs show that
the proposed method improves upon the competing baseline, with  a comparable run time. 




%
%
%
%
%
%
%
%

\section{Unbiased estimator of Schatten-$k$ norms of a graph Laplacian}
\label{sec:schatten}

In this section,  we focus on the unnormalized $L$  as Schatten norms are homogeneous and the normalization can be applied afterwards.
We first provide an alternative method for computing $\|L\|_k$, and show how it leads to
a novel estimator of the Schatten norm from a sampled subgraph.
We use an alternative expression of
the Schatten $k$-norm of a positive semidefinite $L$
as the trace of the $k$-th power:
\begin{eqnarray}
	 (\| L \|_k)^k   \;\; =\;\;  \Tr(L^k)\;.
	 \label{eq:deftrace}
\end{eqnarray}
Such a sum of the diagonal entries is
the sum of weights of all closed walks of length $k$,
where the weight of a walk is defined as follows.
 A length-$k$ closed walk in $G=(V,E)$ is a sequence of vertices
$w=(w_1,w_2,\ldots,w_k,w_{k+1})$ with $w_1=w_{k+1}$ and
either $(w_i,w_{i+1})\in E$ or $w_i=w_{i+1}$ for all $i\in[k]$.
Note that we allow repeated nodes and repeated edges.
Essentially, these are walks in a graph $G$ augmented by self-loops at each of the nodes.
We define the {\em weight} of a walk $w$ in $G$ to be
\begin{eqnarray}
	\mu_G(w)  \;\; \equiv \;\; \prod_{i=1}^k L_{w_iw_{i+1}} \;,
\end{eqnarray}
which is the product  of the weights along the walk and $L=D-A$ is the graph Laplacian.
It follows from \eqref{eq:deftrace} that
\begin{eqnarray}
	\|L\|_k^k \;\; =\;\;  \sum_{w:\text{ all length $k$ closed walks}}  \mu_G(w) \;. \label{eq:defnorm2}
\end{eqnarray}
Even though this formula holds for any general matrix $L$,
it simplifies significantly for graph Laplacians,
as its  all non-zero off-diagonal entries are $-1$ (and its diagonal entries are the degrees of the nodes).
Consider a length-3 walk $w=(u,v,v,u)$ whose pattern is shown in the subgraph $A_2$ in Figure \ref{fig:3cyclic}.
This walk has weight  $\mu_G(w)=(-1)^2 d_v$, where $d_v$ is the degree of node $v$.
Similarly, a walk $(u,u,u,u)$ of pattern $A_1$ in Figure \ref{fig:3cyclic} has
weight $\mu_G(w) = d_u^3$,
and a walk $w=(u,v,x,u)$ of pattern $A_3$ has weight $\mu_G(w) = (-1)^3$.

In general,
for a node $u$ in a walk $w$ of length $k$, let $s_u$ denote the number of self-loops traversed in the walk on node $u$.
Then, it follows that
\begin{eqnarray}
	\mu_G(w) \;\; =\;\;  (-1)^{(k-\sum_{u\in w} s_u)} \prod_{u \in w} d_u^{s_u} \;,
	\label{eq:muG}
\end{eqnarray}
where $d_u$ is the degree of node $u$ in $G$.
 The weight of a walk is $\pm 1$ if there are no self loops in the walk.
 Otherwise,  its absolute value is the product of the degrees of the vertices corresponding to the self loops,
 and its sign is determined by how many non-self loop  edges there are.

The first critical step in our approach is
to partition the summation in Eq.~\eqref{eq:defnorm2}
according to the pattern of the respective walk,
which will make $(i)$ counting those walks of the same pattern more efficient;
and $(ii)$ also de-biasing straight forward (see Equation \eqref{eq:estimate}) under ransom sampling.
We refer to component-wise scaling w.r.t.~the inverse of the probability of being sampled as de-biasing,
which is a critical step in our approach and will be explained in detail later in this section.
Following the notations from enumeration of small cycles in \cite{AYZ97} and \cite{KO17},
we use the family of patterns called {\em $k$-cyclic pseudographs}:
\begin{eqnarray}
	\|L\|_k^k \;\;=\;\; \sum_{H \in \bH_k} \; \sum_{w:\cH(w)=H} \mu_G(w)\;,
	\label{eq:partitionsum}
\end{eqnarray}
where $\bH_k$ is the set of {\em patterns} that have $k$ edges,
and $\{w:\cH(w)=H\}$ is the set of walks on $G$ that have the same pattern $H$.
We give formal definitions below.
$k$-cyclic pseudographs expand the standard notion of simple k-cyclic graphs,
and include multi-edges and loops, which explains the name pseudograph.

\begin{definition}{\rm
Let  $C_k=(V_k,E_k)$  denote the undirected simple cycle  with $k$ nodes.
An unlabelled and undirected pseudograph $H=(V_H,E_H)$ is called
a {\em $k$-cyclic pseudograph} for $k\geq3$ if
there exists an onto node-mapping from $C_k=(V_k,E_k)$, i.e. $f:V_k \to V_H$, and a one-to-one edge-mapping $g:E_k\to E_H$
such that $g(e)=(f(u_e),f(v_e))$ for all $e=(u_e,v_e) \in E_k$.
We use $\bH_k$ to denote the set of all $k$-cyclic pseudographs.
We use   $c(H)$ to
the number of different node mappings $f$ from $C_k$ to a $k$-cyclic pseudograph $H$.
Each closed walk $w$ of length $k$  is associated with one of the graphs in $\bH_k$,
as there is a unique $H$ that the walk is an Eulerian cycle of under a one-to-one mapping of the nodes.
We denote this graph by $\cH(w)\in \bH_k$.
}
\label{def:pseugograph}
\end{definition}

\tikzset{vertex/.style = {shape=circle,fill=black,inner sep=3pt,draw}}
\tikzset{every loop/.style={min distance=10mm,looseness=10}}
\tikzset{me/.style={to path={
\pgfextra{%
 \pgfmathsetmacro{\startf}{-(#1-1)/2}  
 \pgfmathsetmacro{\endf}{-\startf} 
 \pgfmathsetmacro{\stepf}{\startf+1}}
 \ifnum 1=#1 -- (\tikztotarget)  \else
\foreach \i in {\startf,\stepf,...,\endf}
    {%
     (\tikztostart)        parabola[bend pos=0.5, thin] bend +(0,0.3*\i)  (\tikztotarget)
      }
      \fi   
     \tikztonodes
      }}} 

\captionsetup[subfloat]{labelformat=empty}

\begin{figure}[h]
\centering
\begin{tikzpicture}
\def\fixsize{0.7}
\matrix [name=m,
cells={anchor=south},
column sep=0.1cm,row sep = 0.1cm]
{
\setcounter{subfigure}{7}
\node[scale=\fixsize] (a1) {\subcaptionbox{\large{$A_1$}}{\begin{tikzpicture}
\node[vertex] (a) at (-1,0) {};
\draw[thin] (a) edge [in=0,out=60,loop]();
\draw[thin] (a) edge [in=120,out=180,loop]();
\draw[thin] (a) edge [in=240,out=300,loop]();
\end{tikzpicture}} }; &
\node[scale=\fixsize] (a2) {\subcaptionbox{\large{$A_2$}}{\begin{tikzpicture}

\node[vertex] (a) at (-1,0) {};
\node[vertex] (b) at (1,0) {};
\draw[thin] (a) edge[me=2] (b); 

\draw[thin] (b) edge [in=60,out=120,loop]();

\end{tikzpicture}} }; &
\node[scale=\fixsize] (a3) {\subcaptionbox{\large{$A_3$}}{\begin{tikzpicture}

\node[vertex] (a) at (-1,0) {};
\node[vertex] (b) at (1,0) {};
\node[vertex] (c) at (0,1.73) {};
\draw[thin] (a) edge (b); 
\draw[thin] (a) edge (c); 
\draw[thin] (b) edge (c);

\end{tikzpicture}} }; \\
};
\end{tikzpicture}
\put(-170,-8){$c(A_1)=1$}
\put(-110,-8){$c(A_2)=3$}
\put(-50,-8){$c(A_3)=6$}
\caption{The 3-cyclic pseudographs $\bH_3=\{A_1,A_2,A_3\}$.}
\label{fig:3cyclic}
\end{figure}
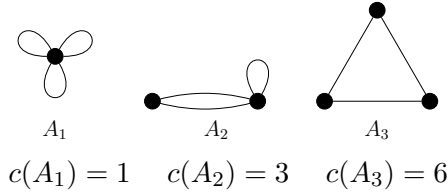

Figure \ref{fig:3cyclic} shows examples of all 3-cyclic pseudographs.
$\bH_3= \{A_1,A_2,A_3\}$ and each one is a distinct pattern
that can be mapped from a triangle graph $C_3$.
In the case of  $A_1$, there is only one mapping from $C_3$ to $A_1$ and
corresponding multiplicity is $c(A_1)=1$.
Also, a walk $w=(u,u,u,u)$ on the graph $G$
 has pattern $A_1$, which we denote by
$\cH(w)=A_1$.
In the case of $A_2$,
any of the three nodes can be mapped to the left-node of $A_2$,
which gives $c(A_2)=3$.
In the case of $A_3$, each permutation of the three nodes are distinct,
which gives $c(A_3)=6$.
We show more examples of length 4 in Figure \ref{fig:4cyclic}.
$k$-cyclic pseudographs for larger $k$ can be enumerated as well  (e.g.~\cite{KO17}).

\tikzset{vertex/.style = {shape=circle,fill=black,inner sep=3pt,draw}}
\tikzset{every loop/.style={min distance=10mm,looseness=10}}
\tikzset{me/.style={to path={
\pgfextra{%
 \pgfmathsetmacro{\startf}{-(#1-1)/2}  
 \pgfmathsetmacro{\endf}{-\startf} 
 \pgfmathsetmacro{\stepf}{\startf+1}}
 \ifnum 1=#1 -- (\tikztotarget)  \else
\foreach \i in {\startf,\stepf,...,\endf}
    {%
     (\tikztostart)        parabola[bend pos=0.5, thin] bend +(0,0.3*\i)  (\tikztotarget)
      }
      \fi   
     \tikztonodes
      }}} 

\captionsetup[subfloat]{labelformat=empty}

%

\begin{figure}[h]
\centering
\begin{tikzpicture}
\def\fixsize{0.7}
\matrix [name=m,
cells={anchor=south},
column sep=0.1cm,row sep = 0.1cm]
{
\setcounter{subfigure}{7}
\node[scale=.8*\fixsize] (b1) {\subcaptionbox{\large{$B_1$}}{\begin{tikzpicture}
\node[vertex] (a) at (-1,0) {};
\draw[thin] (a) edge [in=-30,out=30,loop]();
\draw[thin] (a) edge [in=60,out=120,loop]();
\draw[thin] (a) edge [in=150,out=210,loop]();
\draw[thin] (a) edge [in=240,out=300,loop]();
\end{tikzpicture}} }; &
\node[scale=.8*\fixsize] (b2) {\subcaptionbox{\large{$B_2$}}{\begin{tikzpicture}

\node[vertex] (a) at (-1,0) {};
\node[vertex] (b) at (1,0) {};
\draw[thin] (a) edge[me=4] (b);

\end{tikzpicture}} }; &
\node[scale=.8*\fixsize] (b3) {\subcaptionbox{\large{$B_3$}}{\begin{tikzpicture}

\node[vertex] (a) at (-1,0) {};
\node[vertex] (b) at (1,0) {};

\draw[thin] (a) edge[me=2] (b); 
\draw[thin] (b) edge [in=90,out=30,loop]();
\draw[thin] (b) edge [in=90,out=150,loop]();
\end{tikzpicture}} }; \\
\node[scale=.8*\fixsize] (b4) {\subcaptionbox{\large{$B_4$}}{\begin{tikzpicture}

\node[vertex] (a) at (-1,0) {};
\node[vertex] (b) at (1,0) {};
\node[vertex] (c) at (-3,0) {};

\draw[thin] (a) edge[me=2] (b); 
\draw[thin] (a) edge[me=2] (c); 
\end{tikzpicture}} }; &
\node[scale=.8*\fixsize] (b5) {\subcaptionbox{\large{$B_5$}}{\begin{tikzpicture}

\node[vertex] (a) at (-1,0) {};
\node[vertex] (b) at (1,0) {};
\draw[thin] (a) edge[me=2] (b); 

\draw[thin] (b) edge [in=60,out=120,loop]();
\draw[thin] (a) edge [in=120,out=60,loop]();

\end{tikzpicture}} }; &
\node[scale=.8*\fixsize] (b6) {\subcaptionbox{\large{$B_6$}}{\begin{tikzpicture}

\node[vertex] (a) at (-1,0) {};
\node[vertex] (b) at (1,0) {};
\node[vertex] (c) at (-1,2) {};
\node[vertex] (d) at (1,2) {};
\draw[thin] (a) edge[me=1] (b); 
\draw[thin] (b) edge[me=1] (d);
\draw[thin] (c) edge(a);
\draw[thin] (c) edge (d);

\end{tikzpicture}} }; &
\node[scale=.8*\fixsize] (b7) {\subcaptionbox{\large{$B_7$}}{\begin{tikzpicture}

\node[vertex] (a) at (-1,0) {};
\node[vertex] (b) at (1,0) {};
\node[vertex] (c) at (0,1.73) {};

\draw[thin] (a) edge[me=1] (b); 
\draw[thin] (a) edge(c);
\draw[thin] (b) edge(c);

\draw[thin] (c) edge [in=120,out=60,loop]();

\end{tikzpicture}} };\\
};
\end{tikzpicture}

\caption{The 4-cyclic pseudographs $\bH_4$.}
\label{fig:4cyclic}
\end{figure}
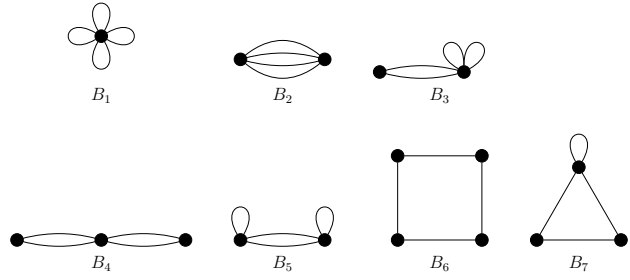



For a pattern $H$, let $S_H$ denote the set of self-loops in $H$,
and $s_u$ denote the number of self loops at node $u$ in the walk $w$.
Then the summation of walks
can be partitioned according to their patterns as:
\begin{eqnarray}
	\label{eq:defnorm3}
	\|L\|_k^k &=& \sum_{H \in \bH_k}  (-1)^{k-|S_H|} \Big\{ \sum_{w:\cH(w)=H}  \prod_{u\in w} d_u^{s_u}  \Big\} \,,
\end{eqnarray}
which follows from substituting \eqref{eq:muG} in \eqref{eq:partitionsum}.
This expression does not require the (computation of) singular values and
leads to a natural unbiased estimator given a sampled subgraph.
As the probability of a  walk being sampled depends only on the pattern,
we introduce a novel estimator $\hTheta_k(G_\Omega)$ of $\|L\|_k^k$ that de-biases each pattern separately:
\begin{eqnarray}
	   \hTheta_k(G_{\Omega})  =
	   \sum_{H \in \bH_k} \frac{(-1)^{k-|S_H|} }{p^{|V_H|}}  \Big\{ \sum_{w: \cH(w)=H} \theta_w(G_\Omega) {\mathbb I}(w\subseteq G_\Omega) \Big\} \;,
	   \label{eq:estimate}
\end{eqnarray}
where $|V_H|$ is the number of {\em nodes} in $H$,
$p^{|V_H|}$ is the probability that walk with pattern $H$ is sampled (i.e.~all edges involves in the walk are present in the
sampled subgraph $G_\Omega$),
and ${\mathbb I}(w\subseteq G_\Omega)$ denotes the indicator that all nodes in the walk $w$ are sampled.  $\theta_w(G_\Omega)$  is defined below.

As the degrees of the nodes in the original graph are unknown,
it is challenging to estimate the polynomial of the degrees $\prod_{u\in w} d_u^{s_u}$
in Eq.~\eqref{eq:defnorm3}, from the sampled graph.
To this end, we introduce a novel estimator $\theta_w(G_\Omega)$ in Section \ref{sec:schatten_theta},
which  is unbiased; it satisfies
$$
\E[\theta_w(G_\Omega) | {\mathbb I}(w \subseteq G_\Omega)] \;\; =\;\; \prod_{u\in w} d_u^{s_u}\;.
$$
It immediately follows by taking the expectation of \eqref{eq:estimate},
that $\hTheta_k(G_\Omega)$ is unbiased, i.e.
\begin{eqnarray}
	\E_\Omega[\hTheta_k(G_\Omega)] \;\; =\;\; \|L\|_k^k\;.
	\label{eq:unbiased}
\end{eqnarray}

\section{An unbiased estimator of the polynomial of the degrees}
\label{sec:schatten_theta}

\begin{figure}[h]
	\begin{center}
	\includegraphics[width=.45\textwidth]{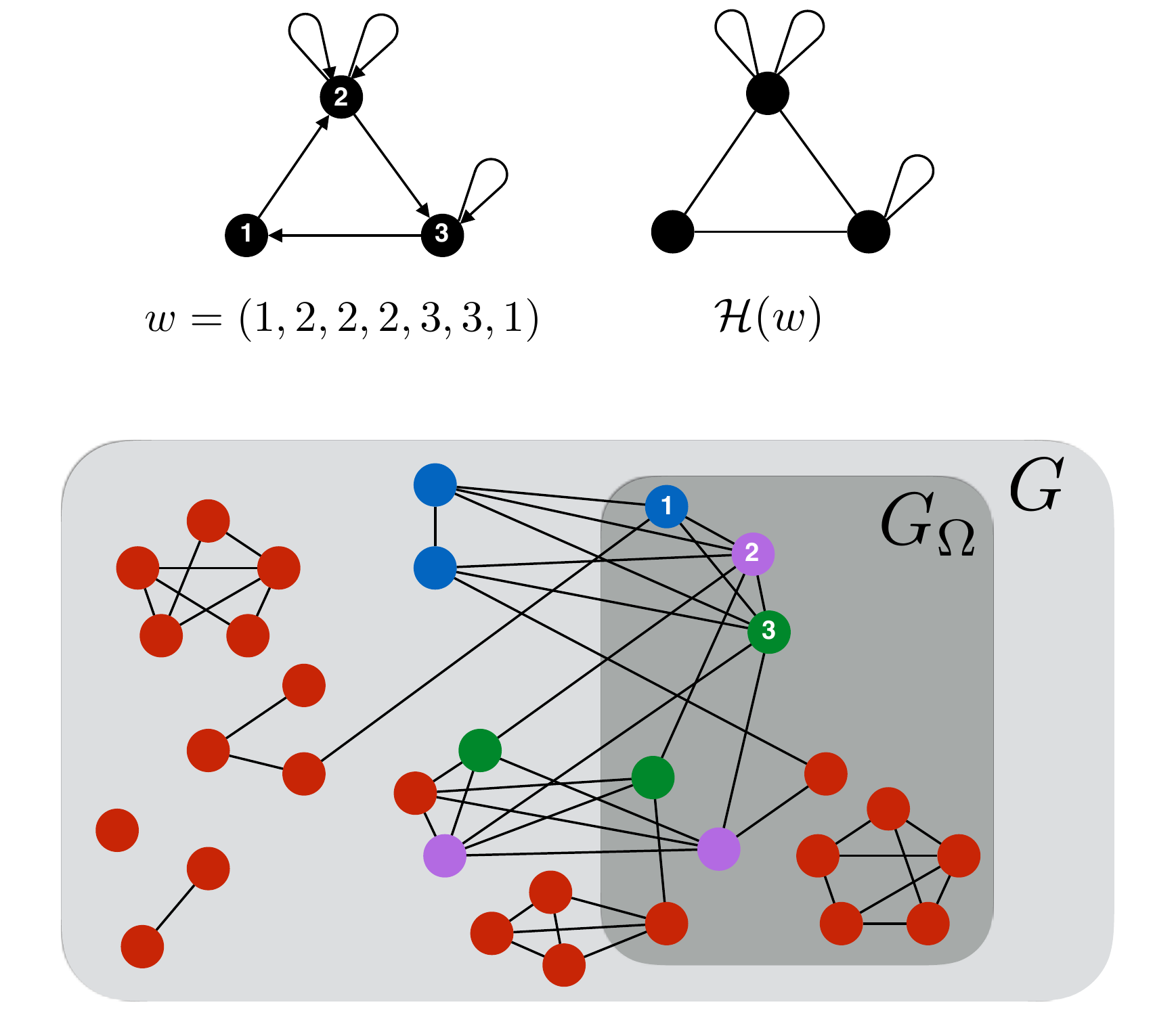}
	\end{center}
	\caption{
		We are partitioning the original graph $G$ with respect to
		a length-$(k=6)$ closed walk $w=(1,2,2,2,3,3,1)$.
		Its corresponding $k$-cyclic pseudograph $\cH(w) \in \bH_6$ is shown on the top.
		Red nodes are not connected to either $2$ or $3$, which are the nodes of interest in $w$ as they have self loops. Blue nodes are only connected to $2$, purple to only $3$, and
		blue to both $2$ and $3$.}
	\label{fig:partition}
\end{figure}

Our strategy to get an unbiased estimator  of $\prod_{u\in w} d_u^{s_u}$
is to  first  partitioning the nodes in the original graph $G$ to get
a more insightful factorization of $\prod_{u\in w} d_u^{s_u}$ in Eq.~\eqref{eq:sumdegree2} (see Figure \ref{fig:partition})
that removes dependences between the summands,
and next by estimating each term independently in the factorization.

Consider a concrete task of
estimating
$\prod_{u\in w} d_u^{s_u} = (d_2)^2 (d_3)^1 = 6^2\times 6$,
for a walk $w=(1,2,2,2,3,3,1)$ in the observed subgraph $G_\Omega$.
Note that we only see $G_\Omega$, whose degrees are very different from the original graph.
For instance, node 2 now has degree 3 and node 3 has degree 3 in the sampled graph.
Further, these random variables (the observed degrees) are correlated, making estimation challenging.
To make such correlations apparent, we first give a
novel  partitioning of the nodes $V$ in the following.

\subsection{Partitioning $V$}

Our strategy is first to
partition the nodes $V$ in the original graph,
with respect to a walk $w=(w_1,\ldots,w_{k+1})$ of interest.
For a closed walk $w$, let $U=\{u_1,\ldots,u_\ell\}$ denote the set of nodes in $w$ that have at least one self-loop,
 let $\ell=|U|$ denote its cardinality, and  let $\{s_1,\ldots,s_\ell\}$ denote the number of self-loops at each node.
In the running example, we have
$U=\{u_1=2,u_2=3\}$, $\ell=2$, $s_1=2$, and $s_2=1$.
As our goal is to estimate $(d_2)^2(d_3)$,
we partition the nodes with respect to how they relate to the nodes in $U=\{2,3\}$.
Concretely,
there are four partitions:
nodes that are not connected to either $2$ or $3$ (shown in red in Figure \ref{fig:partition}),
nodes that are only connected to $2$ (shown in green),
nodes that are only connected to $3$ (shown in purple),
and nodes that are connected to both $2$ and $3$ (shown in blue).
Nodes in each partition contribute in different ways to the target quantity
$(d_2)^2(d_3)$, which will be precisely captured in the {\em factorization} in Eq.~\eqref{eq:sumdegree2}.
In general, we need to consider all such variations in the partitioning, which gives
\begin{eqnarray}
	V \;\; = \;\; \bigcup_{T \subseteq U} V_{T,U\setminus T} \;, \label{eq:partitions}
\end{eqnarray}
where $V_{T,T'}=\big\{  \bigcap_{v\in T} \partial v\big\} \bigcap \big\{ \bigcap_{v\in T'} \partial v^c \big\}$ is the set of nodes that are adjacent to all nodes in $T$ but are not adjacent to any nodes in $T'$,
and  $\partial v$ denotes the neighborhood of node $v$ and $\partial v^c$ denotes the
complement of $\partial v$.
We let $V_{\emptyset,U} = \bigcap_{v\in U} \partial v^c$ and $V_{U,\emptyset} = \bigcap_{v\in U} \partial v$.
Essentially, we are labelling each node according to which nodes in $U$ it is adjacent to, and grouping those nodes with the same label.
In the running example,
$V = V_{U,\emptyset} \cup V_{\{3\},\{2\}} \cup V_{\{2\},\{3\}} \cup V_{\emptyset,U} $,
where the partitions are subset of nodes in
blue, purple, green, and red, respectively.

Let $d_{T,U\setminus T}=|V_{T,U\setminus T}|$ denote the size of a partition such that
\begin{eqnarray}
	d_u \; \; =\; \; \sum_{ T \in \cT_u }  d_{T,U\setminus T}  \;,
\end{eqnarray}
for any $u\in U$  where $\cT_u=\{ T \subseteq U | u\in T \} $ is the set of subsets of $U$ containing $u$.
For example, $d_2=6$ which is the sum of blue and green nodes,
and $d_3=6$ which is the sum of blue and purple nodes.

We are partitioning the neighborhood of $u$ such that each term can be separately estimated. 
This ensures  we handle the correlations among the degrees of different nodes in $w$ correctly.
The quantity of interest  is
\begin{align}
	& \prod_{i \in [\ell]} d_{u_i}^{s_i}  =  \prod_{i\in [\ell]} \Big( \sum_{ T \in \cT_{u_i}  }  d_{T,U\setminus T }  \Big)^{s_i}
	\nonumber\\
		& =  \sum_{\big( T_{1}^{(1)},\ldots, T_{1}^{(s_{1})} , \cdots , T_{\ell}^{(1)},  \cdots , T_{\ell}^{(s_{\ell})}  \big) \in  (\cT_{u_1})^{s_{1}}\times \cdots \times (\cT_{u_\ell})^{s_{\ell}} }
		\Big\{ \prod_{j=1}^{\ell} \prod_{i=1}^{s_{j}} d_{T_{j}^{(i)},U\setminus T_{j}^{(i)} }   \Big\} \;,
		\label{eq:sumdegree2}
\end{align}
where $T_j^{(i)}$ is a $i$-th choice of a set in $\cT_{u_j}$ that contains the node $u_j$
for $i\in[s_j]$, and $[\ell]=\{1,\ldots,\ell\}$ denotes the set of positive  integers up to $\ell$.
The second equation follows directly from exchanging the product and the summation.
This alternative expression is crucial in designing an unbiased estimator,
since each term in the summation can now be estimated separately as follows.

Consider a task of estimating a  single term in \eqref{eq:sumdegree2},
and we merge those $T_j^{(i)}$'s that happen to be identical:
\begin{eqnarray}
	\prod_{j=1}^{\ell} \prod_{i=1}^{s_{j}} d_{T_{j}^{(i)},U\setminus T_{j}^{(i)} } \;\; = \;\;
		\prod_{T \in \bT}  (d_{T,U\setminus T })^{t_T}\; ,\label{eq:sumdegree3}
\end{eqnarray}
where $\bT=\{T_{1}^{(1)},\ldots, T_{1}^{(s_{1})} , \cdots , T_{\ell}^{(1)},  \cdots , T_{\ell}^{(s_{\ell})} \}$ is the current set of partitions allowing for multiple entries of the same set,
and $t_T$  is the multiplicity, i.e.~how many times a set $T$ appears in the set $\bT =$
$(T_{1}^{(1)},\ldots, T_{1}^{(s_{1})} , \cdots , T_{\ell}^{(1)},  \cdots , T_{\ell}^{(s_{\ell})} )$.
Each term in the right-hand side can be now separately estimated,
as $(a)$ $V_{T,U\setminus T}$'s are disjoint and
$(b)$ we know for the sampled subgraph the membership of each sampled node.
This follows from the fact that,
conditioned on the event that $\{w\subseteq \Omega\}$,
we know how the sampled nodes in $\Omega$ are connected to  any node in $\{w_i\}_{i=1}^{k+1}$
and in particular those with self-loops denoted by $U$.
Hence, for any node in $\Omega$ the membership  (or the  color in the Figure \ref{fig:partition}) is trivially revealed.
Therefore, we can handle (the degrees $d_{T,U\setminus T}$ in) each partition separately, and
estimate each monomial in $\prod_{T \in \bT}  (d_{T,U\setminus T })^{t_T}$.
The problem is reduced to the task of estimating $d_{T,U\setminus T}^s$ for some integer $s$
and some partition $V_{T,U\setminus T}$.


\subsection{Unbiased estimator of $d_{T,U\setminus T}^s$}
From the original graph $G=(V,E)$ (where the size of each partition is denoted by $d_{T,U\setminus T}$),
we observe a sampled subgraph $G_\Omega=(\Omega,E_\Omega)$
(where the size of each partition in $G_\Omega$ is denoted by $d_{T,U \setminus T}(\Omega)$),
and we let  $d_{T,U\setminus T}(w)$ denote the size of the partition intersecting the walk $w=(w_1,\ldots,w_{k+1})$.
Precisely, $d_{T,U\setminus T}(\Omega) \equiv |V_{T,U\setminus T} \bigcap \Omega|$,
and $d_{T,U\setminus T}(w) \equiv |V_{T,U\setminus T} \bigcap \{w_i\}_{i=1}^{k+1}|$.
We do not allow multiple counts when computing the size, such that
$d_{\{2\},\{3\}}(\Omega)=2$
and $d_{\{2\},\{3\}}(w)=1$, in the example.

Let us focus on a particular walk $w$ on a graph $G$,
its corresponding $U$ and  a fixed $T\subseteq U$, such that $V_{T,U\setminus T}$ and $d_{T,U\setminus T}$ are fixed.
Now $d_{T,U\setminus T}(\Omega)$ is a  random variable
 representing  how many nodes in the partition
$V_{T,U\setminus T}$ are sampled.
Conditioned on the fact that $w$ is sampled, and hence
a $d_{T,U\setminus T}(w)$ sampled nodes are already observed,
the remaining $(d_{T,U\setminus T}-d_{T,U\setminus T}(w))$ nodes
are sampled i.i.d.~with  probability $p$.
Hence, conditioned on $\{w\subseteq \Omega\}$, the size of the sampled partition is distributed as
 \begin{eqnarray}
 	d_{T,U\setminus T}(\Omega) \sim {\rm Binom}(d_{T,U\setminus T}-d_{T,U\setminus T}(w),p) + d_{T,U\setminus T}(w)\;.
\end{eqnarray}
This leads to a natural unbiased estimator of the monomial $d_{T,U\setminus T}^s$ as
\begin{eqnarray}
	\widehat{d}_{T,U\setminus T}^{(s)}  \;\; = \;\;  \langle ( \, A^{-1})_{s+1} \, ,\,  \overline{d} \,  \rangle\;,
	\label{eq:estmate_monomial}
\end{eqnarray}
where
$\overline{d}=[1\,,\,d_{T,U\setminus T}(\Omega)\,,\,d_{T,U\setminus T}(\Omega)^2,\ldots,d_{T,U\setminus T}(\Omega)^s ]^\top$ is a column vector in $\reals^{s+1}$ of the monomials of the observed size of the partition,
$A$ is the unique matrix satisfying
\begin{eqnarray}
	\E[\overline{d}] \;\; = \;\; A \, [1\,,\,   d_{T,U\setminus T} \,,\,\ldots\,,\, d_{T,U\setminus T}^s ]^\top \;,
\end{eqnarray}
and $(A^{-1})_{s+1}$ is the $(s+1)$-th row of $A^{-1}$.
One can check immediately that
$\E[\widehat{d}_{T,U\setminus T}^{(s)} ] =  \langle ( \, A^{-1})_{s+1}  ,  \E[ \overline{d}] \,  \rangle
= d_{T,U\setminus T}^{s} $, hence giving the desired unbiased estimator.
The matrix $A$ is a lower triangular matrix which depends only on
$s$, $p$ and the structure of the walk via $d_{T,U\setminus T}(w)$.
In terms of these three parameters,
the required vector $(A^{-1})_{s+1}$ has a closed form expression,
and hence the estimator can be computed in a straight forward manner.
It uses the moments of a binomial distribution, which can be computed immediately.

An example of $A$ for $s=3$ is given in \eqref{eq:defA},
where one should plug-in
$\ell= d_{T,U\setminus T}(w)+1$, $\wgt = d_{T,U\setminus T}+1$ and $\widetilde\tau = d_{T,U\setminus T}(\Omega)$.
This leads to an unbiased estimator of $\prod_{i \in [\ell]} d_{u_i}^{s_i}$ by
replacing \eqref{eq:estmate_monomial} into \eqref{eq:sumdegree3} and \eqref{eq:sumdegree2}:
\begin{align}
&	\theta(w,G_\Omega) \;\; = \;\;  \nonumber\\
&	 \sum_{\big( T_{1}^{(1)},\ldots, T_{1}^{(s_{1})} , \cdots , T_{\ell}^{(1)},  \cdots , T_{\ell}^{(s_{\ell})}  \big) \in  (\cT_{u_1})^{s_{1}}\times \cdots \times (\cT_{u_\ell})^{s_{\ell}} }
		\Big\{
		\prod_{T\in \bT}      \widehat{d}_{T,U\setminus T}^{(t_T)}
		\Big\}
		\;. \label{eq:estimate_theta}
\end{align}
By construction, it is immediate that the estimator is unbiased:
$\E[ \theta(w,G_\Omega) | I (w\subseteq \Omega)] = \prod_{u \in w} d_{u}^{s_u}$.



\section{Polynomial approximation}
\label{sec:polynomial}

The remaining goal in our approach is to
design a polynomial approximation of the target function
$H_\alpha :[0,1]\to[0,1]$ defined in  \eqref{eq:eq3} for a fixed scalar $\alpha \in (0,1)$.
Concretely,
for a given integer $m$,
we want a degree-$m$ polynomial approximation $f(x)$ of $H_\alpha(x)$
such that
$(i)$ $f(0)=0$;
$(ii)$ the approximation error (as measured by the $\ell_\infty$ norm)
is small in the interval $[\alpha,1]$;
and $(iii)$ we can provide an upper bound on the approximation error:
$\max_{x\in[\alpha,1]} |H_\alpha(x)-f(x)|$.
The first condition can be met by any function with proper scaling and shifting,
and strictly enforcing it ensures that we make fair comparisons.
The second condition ensures we have a good approximation, as
the non-zero singular values only lie in the interval $[\alpha,1]$.
In particular, the approximation error outside of this interval is irrelevant.
The last condition ensures we get the desired performance guarantees for
the estimation error of the number of connected components.
The (upper bound on the)
approximation error of the polynomial function directly translates into the end-to-end
error on the estimation.

\begin{figure}[h]
	\begin{center}
	\includegraphics[width=.48\textwidth]{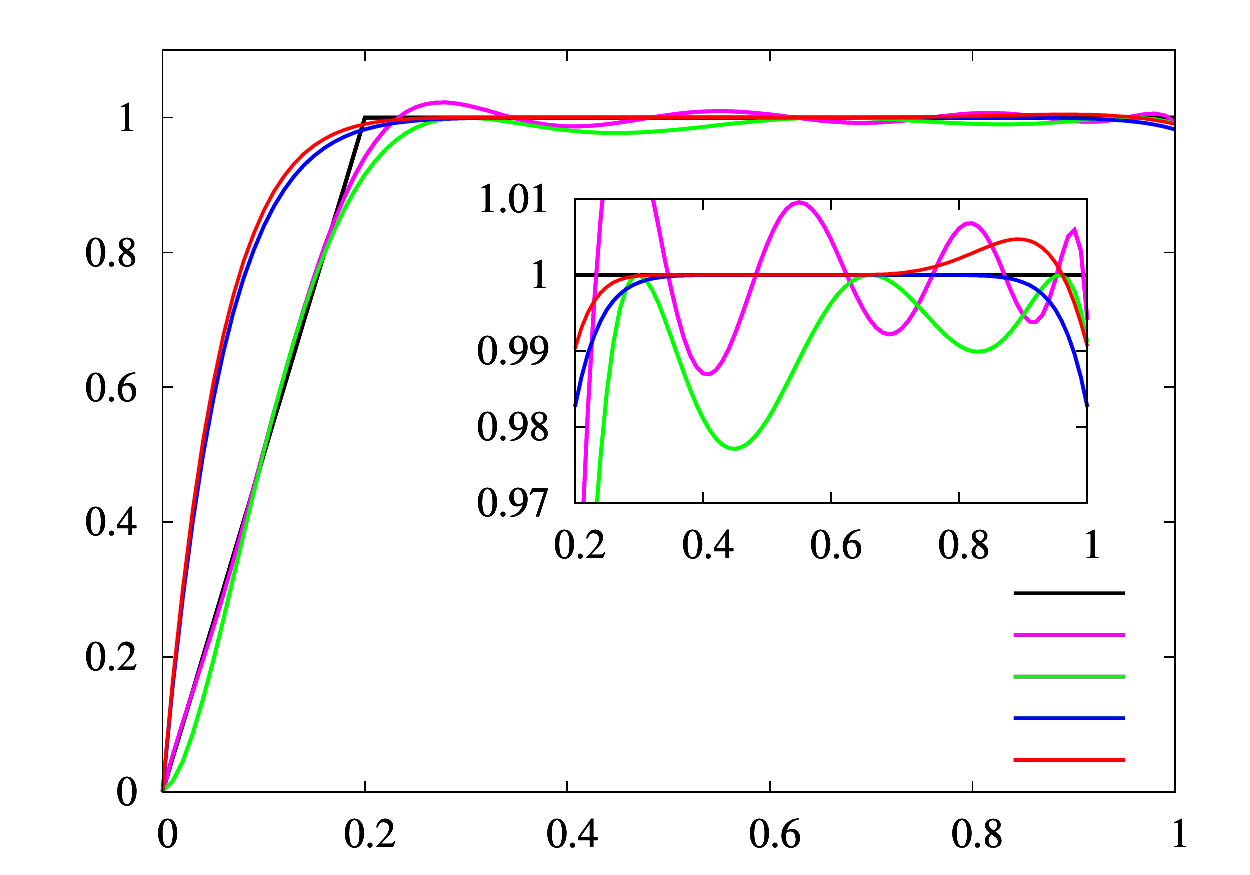}
	\put(-175,170){$H_\alpha(x)$ and its approximations}
	\put(-115,-5){$x$}
	\put(-68,54){\tiny $H_{0.2}(x)$}
	\put(-96,45){\tiny Chebyshev appx. $C(x)$}
	\put(-96,37){\tiny Composite appx. $B(x)$}
	\put(-62,30){\tiny $f_{\tb}(x)$}
	\put(-62,22){\tiny $f_{\hb}(x)$}
	\\
	\includegraphics[width=0.48\textwidth]{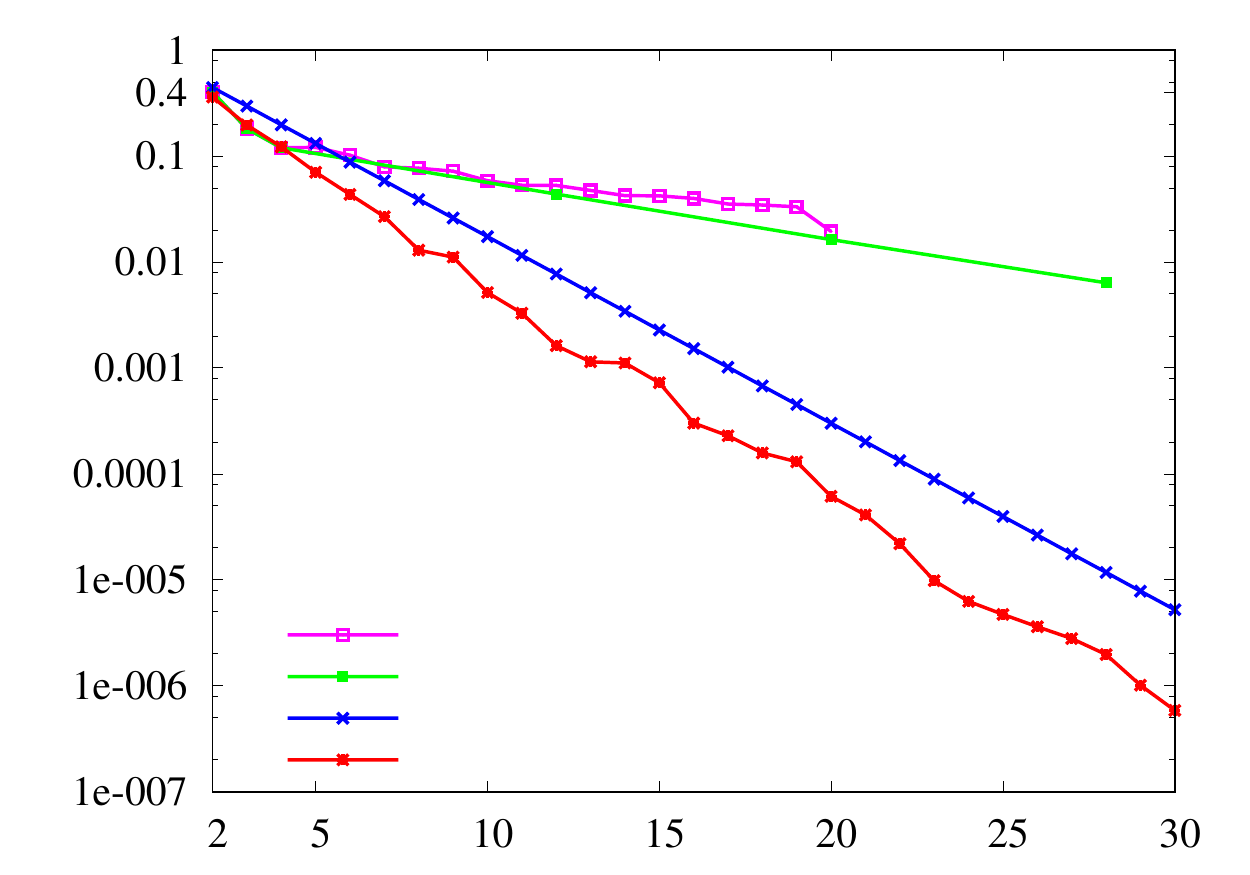}
	\put(-230,167){$\ell_\infty$ Error in $[\alpha,1]$}
	\put(-190,-5){degree $\mmm$ of the polynomial aproximations}
	\put(-163,22){\tiny $f_{\hb}(x)$}
	\put(-163,30){\tiny $f_{\tb}(x)$}
	\put(-163,37){\tiny Composite appx. $B(x)$}
	\put(-163,46){\tiny Chebyshev appx. $C(x)$}
	\end{center}
	\caption{Top: four polynomial approximations
	to $H_{0.2}(x)$ of degree $k=10$,
	$f_{\hb}(x)$ with $\hb$ chosen according to Algorithm \ref{alg:poly},
	$f_{\tb}(x)$ with $\tb=(2/1.2){\mathbf 1}$ as prescribed above,
	Chebyshev approximation $C(x)$, and
	the composite approximation $B(x)$ from \cite{ZWJ15}.
	Bottom: approximation error  achieved by the proposed
	$f_{\hb}(x)$  improves upon
	other  polynomial functions.
	}
	\label{fig:function}
\end{figure}
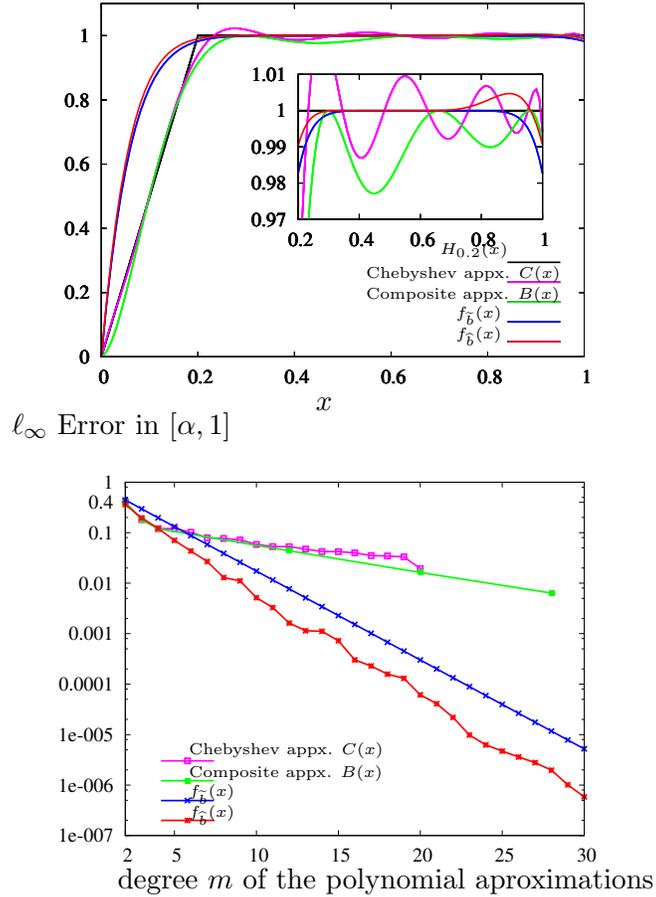

A first attempt might be to use a
Chebyshev approximation \cite{Che1853,MH02} directly on $H_\alpha(x)$.
 This is  optimal in terms of achieving a target $\ell_\infty$ error
 with the smallest degree in all regimes of $[0,1]$.
 However, we only care about the $\ell_\infty$ error in $[\alpha,1]$.
 As shown in magenta curve in
Figure \ref{fig:function},
the Chebyshev approximation $C(x)$ unnecessarily fits the curve in
$(0,\alpha]$, resulting in larger error in $[\alpha,1]$.

A natural fix is to use
filter design techniques, e.g.~Parks-McClellan algorithm \cite{PM72},
where Chebyshev polynomials have been applied to design
high pass filters  with similar constraints as ours.
This will give a polynomial approximation with
small approximation error in the desired pass band of $[\alpha,1]$.
However, these techniques do not come with the desired approximation guarantee
that we seek.

One approach proposed in \cite{ZWJ15}
does come with a provable error bound.
This approximation $B(x)$ composes  a Chebyshev approximation of a constant degree
$q$ with
the CDF of a beta distribution of degree $(m/q-1)/2$.
The beta distribution boosts the approximation of the function in the interval $[\alpha,1]$,
thus providing an error bound of $O((c_\alpha)^m)$, where $c_\alpha$ is a constant that depends on $\alpha$.
Figure~\ref{fig:function} shows that $B(x)$ (in green) still unnecessarily
fits the curve in $(0,\alpha]$, as it starts with a
(lower-degree) Chebyshev approximation of $H_\alpha(x)$.

Our goal is to design a new polynomial approximation
that ignores the region $(0,\alpha]$ completely,
such that it achieves improved performance in $[\alpha,1]$,
and also comes with a provable error bound.
We propose using a parametric family that ensures $f_b(0)=0$:
\begin{eqnarray}
	f_\bb(x)  \;\; =\;\; 1- \prod_{i=1}^\mmm (1-\bb_i x)\;, \label{eq:defpoly}
\end{eqnarray}
for a vector
$\bb=[\bb_1 , \ldots , \bb_\mmm] \in \reals^\mmm$.
We provide an upper bound on the
approximation error achieved by  the optimal $\bb^*$,
provide a choice of $\tb$ in a closed form that
achieves the same error bound, and provide a heuristic for locally searching for the optimal $\bb^*$
to improve upon the closed-form $\tb$.

\begin{propo}
	\label{pro:poly}
	For any $\alpha\in(0,1)$ and $\mmm\geq 2$, the optimal parameter $\bb^*  \in \arg \min_{\bb\in\reals^\mmm} \max_{x\in[\alpha,1]} |H_\alpha(x) - f_{\bb}(x)| $ achieves error bounded by
	\begin{eqnarray}
	 	\max_{x\in[\alpha,1]} |H_\alpha(x) - f_{\bb^*}(x)| \;\; \leq \;\; \Big( \frac{1-\alpha}{1+\alpha}\Big)^\mmm \;.
		\label{eq:polyappx}
	\end{eqnarray}
\end{propo}
A proof is provided in Section \ref{sec:poly_proof}.
As the optimal $\bb^*$ is challenging to find, one option is to
simplify the optimization by searching over a smaller space.
By constraining all $\bb_i$'s to be the same,
solving for minimum $\ell_\infty$ error gives a closed form solution
$\tb\equiv (2/(1+\alpha))[1,\ldots,1]$ that achieves the bound in
\eqref{eq:polyappx} with equality, i.e.
$\max_{x\in[\alpha,1]} |H_\alpha(x) - f_{\tb}(x)|  = ( (1-\alpha)/(1+\alpha) )^\mmm $.


For practical use, we prescribe a slightly better approximation using
a local search algorithm in Algorithm \ref{alg:poly}.
The approximation guarantee is compared
 for $\alpha=0.2$ and varying $\mmm$ in
Figure \ref{fig:function} against
the analytical choice $f_{\tb}(x)$,
the standard Chebyshev approximation $C(x)$ of the first kind,
and the approximation $B(x)$ from \cite{ZWJ15}. 
The proposed $f_{\hb}(x)$ significantly improves upon both,
achieving a faster convergence.
The key idea is to exploit the fact that
we care about approximating only in the regime of $[\alpha,1]$.
 There might be other techniques to design better polynomial approximation
than ours, e.g.~\cite{PM72}, but might not come with a performance guarantee.

The inset  in the top panel of Figure \ref{fig:function} illustrates how
the proposed
$\hb$ in red admits more fluctuations  to achieve smaller $\ell_\infty$ error,
compared to the uniform choice of $\tb$.
In Algorithm \ref{alg:poly}, starting from a moderate perturbation around  $\tb$,
we iteratively identify the point $x'$ achieving the maximum error
and update $\bb$ such that the error at $x'$ is decreased.
This  approximation can be done offline for many random initializations for the desired  $\alpha$ and $m$; the one with minimum error can be stored for later use.

\begin{algorithm}                      
\caption{Local search for a polynomial approximation }          
\label{alg:poly}                           
\begin{algorithmic}                    
    \REQUIRE degree $\mmm$, $\alpha$ , number of iterations $T$, step size $\delta>0$
    \ENSURE $\hb \in\reals^\mmm $
    \STATE $\bb_i \Leftarrow (2/(1+\alpha))$ + U$[-1,1]$ for all $i\in[\mmm]$
	\FOR{$t=1$ \TO $T$ }
		\STATE {$x' \Leftarrow \arg\max_{x\in[\alpha,1]} \big| \prod_{i\in[\mmm]}(1-\bb_i x)  \big| $ }
		\STATE {$\bb \Leftarrow \bb + {\rm sign}(1-f_\bb(x') ) \times \delta \times \nabla_\bb f_\bb(x')$ }
	\ENDFOR
\end{algorithmic}
\end{algorithm}

\def\wgtmax{{\wgt_{\rm max}}}
\def\wgtmin{{\wgt_{\rm min}}}

\section{Main results}
\label{sec:main}

The polynomial approximation $f_{\hb}(x)$ of the form \eqref{eq:defpoly}
can easily be translated into the standard polynomial with coefficients $\aa=(\aa_1,\ldots,\aa_\mmm)$ such that
$f_{\hb}(x)=  \aa_1x +  \cdots + \aa_k x^k$.
Together with the Schatten norm estimator
$\hTheta_k(G_{\Omega})$ in \eqref{eq:estimate},
this gives the proposed estimate $\widehat{\cc}(G_\Omega,\alpha,\beta,m)$ in \eqref{eq:eq7}.
We first give an upper bound on the multiplicative error for
a special case of union of cliques, and give a general bound in Theorem \ref{thm:main_ER}.
The overall procedure achieves the following,
 for a special case of {\em union of cliques}, which are also called {\em transitive graphs}:
\begin{thm} If the underlying graph $G$ is a disjoint union of cliques with clique sizes $\wgt_i$, for each connected component
$1 \leq i \leq \cc(G)$,  $\wgt_{\rm max} \equiv \max_{i}\{\wgt_i\}$ and $\wgt_{\rm min} \equiv \min_{i}\{\wgt_i\}$, then for any choice of $\beta \geq \wgtmax$ and $\alpha \leq \wgtmin/\beta$, and any integer $m \geq 1 $,
there exist a function  $g(m) = O(m!)$  and a constant $C>0$  such that for  $\wgt_{\rm min}>C$,
	\begin{align}
		 & \frac{ \E\big[ ( \widehat{\cc}(G_\Omega,\alpha,\beta,m) - \cc(G) )^2 \big] }{\cc(G)^2} \;\; \leq \nonumber \\
		  &  \;\;
	\frac{g(m)\,(1-p^m)}{\cc(G)^2\, p\, \beta^{2}} \sum_{i=1}^{\cc(G)}\bigg( \wgt_{i}^{4} \Big(1 + (\wgt_i p)^{1-2m} \Big) \bigg)   + \gamma^{2m} \frac{N^2}{\cc(G)^2},
		\label{eq:main1}
	\end{align}
	where $\gamma= ({1-\alpha})/({1+\alpha})$.
 Moreover, if there exist  some  positive constants $c_i$'s such that
 $\wgt_i^3 p \geq c_i m!$ or $\wgt_i p^3 \leq c_i/m!$ for all $i$,
 then \eqref{eq:main1} holds with $g(m) = O(c^m)$ for some constant $c>0$.
	\label{thm:main}
\end{thm}

A proof  of Theorem \ref{thm:main} is provided in a longer version of this paper. 
This clearly shows the tradeoff between
the variance (the first term in the RHS) and the bias (the second term in the RHS).
If we choose larger $m$, our functional approximation becomes
more accurate resulting in
a smaller bias.
However, this will require
counting larger patterns
in the estimate of $\hTheta_k(G_\Omega)$,
leading to a larger variance.

In general, the complexity of
our estimator for union of cliques is $O( m \times \cc(G))$, as all relevant quantities to compute
$\hTheta_k(G_\Omega)$ can be pre-computed and
stored in a table for all combinations of $k$ and the size of the observed cliques.
At execution time, we only need to look up one number for each clique we observe
and for each $\hTheta_k(G_\Omega)$ we are estimating.
Hence, the above guarantee also characterizes the trade-off between the computational complexity and
the accuracy.
For example, when spectral gap $\alpha$ is small, we need large $m$ with longer run-time to get bias as small as we need.
We emphasize here that our estimator is generic and
does not assume the true graph is union of cliques.
The same generic estimator happens to be more efficient, when the {\em observed} subgraph is a union of cliques. 

Consider the bias term, which
 captures how the error increases for graphs with smaller spectral gap in $L$.
The normalized spectral gap for union of cliques is $\wgtmin/\wgtmax$,
and balanced components result in a
small spectral gap and a more accurate estimation.

 Consider the variance term, and
as an extreme example, consider the case when all cliques are of the same size $\wgt=N/\cc(G)$.
 It immediately follows that for $\beta=\wgt_{\rm max}=\wgt_{\rm min}$ and $\alpha=1$,
 there is no bias and $\gamma=0$.
 Further assuming $\wgt p >1$, we can choose some small $m=O(1)$ to minimize the variance
 which scales as $O(N^2 /(\cc(G)^3\,p))$.
 Hence, to achieve arbitrarily small error,
 it is sufficient to have sample size $Np$ scale as $(N/\cc(G))^3$.
 This implies that  finite multiplicative error is guaranteed
 only for $\cc(G) = \Omega(N^{2/3})$.

  Such a condition on $\cc(G)$ increasing with respect to $N$
  seems to be unavoidable in general.
  Consider a case when $\cc(G)=c N$ for some constant $c$.
  Then, we need $m=(1/2)\log_\gamma(\delta c^2/2)$
  to make the bias as small as we want,
  say $\delta/2$.
  Suppose the connected components are balanced such that
  $\wgtmax=O(1)$, then
  the variance term will be at most $\delta/2$,
  if $p=\Omega(m^\varepsilon/N)$, where
  $\varepsilon$ depends on $\gamma$ and $c$.

 Note that the best known guarantees for
 estimators tailored for union of cliques
 still require $\cc(G) = \Omega(N^{1-\varepsilon})$ for small
 but strictly positive $p$,
 where the $\varepsilon$ can be made arbitrarily small with
 a small sampling probability $p$
 (e.g.~\cite{Fra78,KW17}).

\begin{figure}[h]
\centering
    \includegraphics[width=0.4\textwidth]{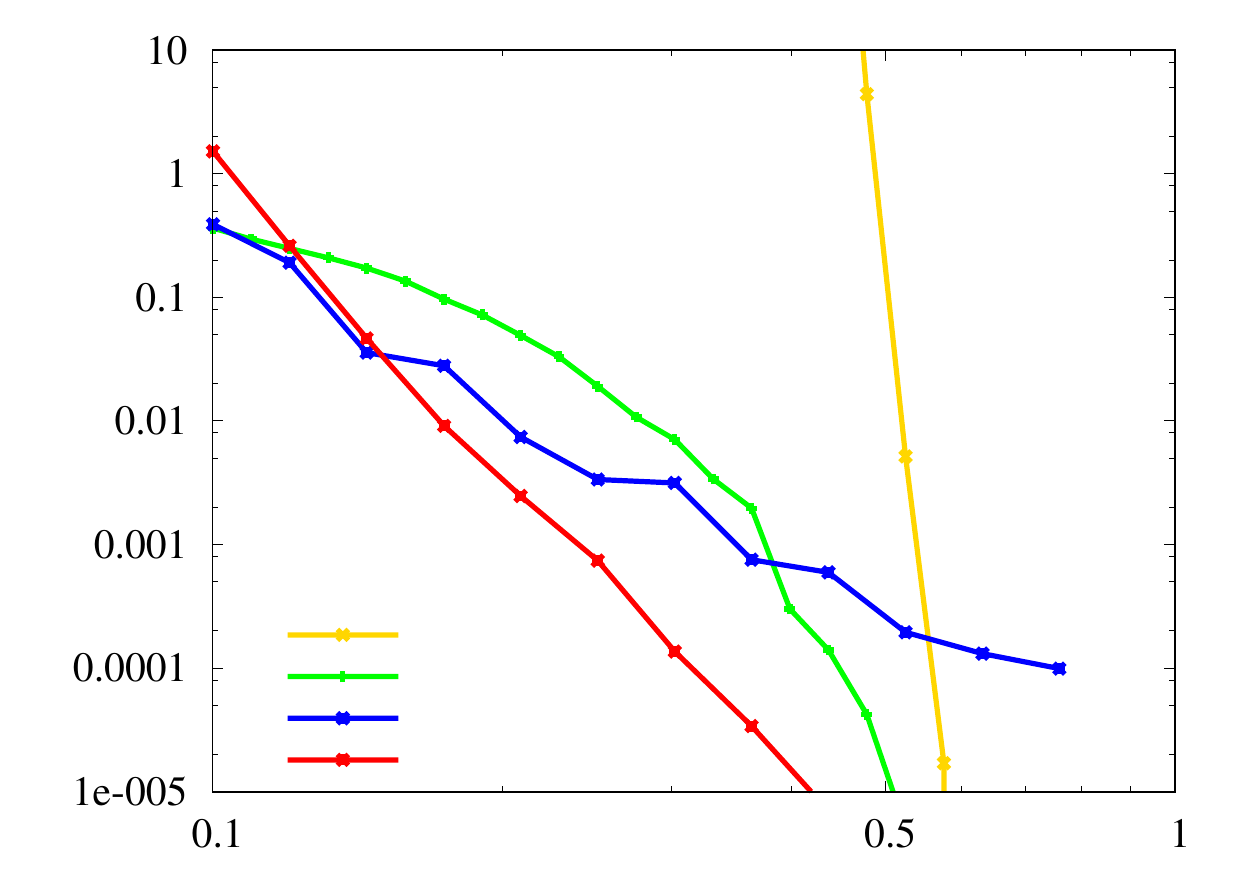}
    	\put(-140,-5){vertex sampling probability $p$}
	\put(-210,35){ \rotatebox{90}{$\frac{\E[ |\widehat{\cc}(G_\Omega,\alpha,\beta,m)-\cc(G)|^2 ]^{1/2}}{|\cc(G)|} $}}
        \put(-134,37){{\tiny $\hcc_{\rm chordal}(G_\Omega)$ }}
        \put(-134,30){{\tiny $\hcc_{\rm clique}(G_\Omega)$ }}
         \put(-134,23){{\tiny  $ \hcc(G_\Omega, 0.5 ,100,10)$ }}
         \put(-134,16){{\tiny  $ \hcc(G_\Omega, 0.5 ,100,15)$ }}
         \hspace{0.4cm}
	\caption{The proposed estimator for two choices of the degree $m$ of the polynomial. 	With the right choice of $m$, we improve upon competing estimators
	 when the original graph is a union of cliques.}
	\label{fig:main_clique}
\end{figure}
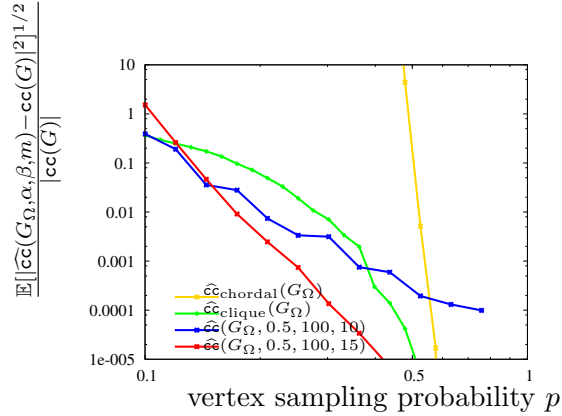

We run synthetic experiments on a graph of $N=3775$ nodes
and union of 50 cliques, each of size $\{51,52,\ldots,100\}$.
Figure \ref{fig:main_clique} shows that we improve upon
three competing estimators for a broad range of $p$.
$\hcc_{\rm chordal}$ is
the best known estimator for chordal graphs from \cite{KW17}, and
$\hcc_{\rm clique}$ is a smoothed version of $\hcc_{\rm chordal}$
explicitly using the knowledge that the underlying graph is a union of cliques.
These  are tailored for chordal graphs and cliques, respectively, and cannot  be
applied to general graphs.
Our generic algorithm, with an appropriate choices of $\alpha$, $\beta$, and $m$,
outperforms these approaches for unions of cliques.
In particular, when $p$ is small, variance dominates and choosing small $m$ helps,
whereas when $p$ is large, bias dominates and choosing large $m$ helps.


\def\dmax{{d_{\rm max}}}
\def\dmin{{d_{\rm min}}}
\begin{thm} For any graph $G$ with size of connected components $\wgt_i$, for
each component $1 \leq i \leq \cc(G)$,  and degree of each node $d_j^{(i)}$, for $1 \leq j \leq \wgt_i$ with $d_{i} \equiv \max_{j}\{d_j^{(i)}\}$, and $\dmax \equiv \max_{i,j}\{d_j^{(i)}\}$, $\dmin \equiv \min_{i,j}\{d_j^{(i)}\}$, for any choice of $\beta \geq 2\dmax$ and $\alpha \leq \sigma_{\rm min}(L)/\beta$, and any integer $m \geq 1 $,
there exist a function  $g(m) = O(2^{m^2})$  such that,
	\begin{align}
		 & \frac{ \E\big[ ( \widehat{\cc}(G_\Omega,\alpha,\beta,m) - \cc(G) )^2 \big] }{\cc(G)^2} \leq   \nonumber \\
		  &
	\frac{g(m)\,(1-p^m)}{\cc(G)^2\, p\, \beta^{2}} \sum_{i=1}^{\cc(G)}\bigg( \wgt_{i}^{2}d_i^2 \Big(1 + (d_i p)^{1-2m} \Big) \bigg)   + \gamma^{2m} \frac{N^2}{\cc(G)^2 },
		\label{eq:main1ER}
	\end{align}
	where $\gamma= ({1-\alpha})/({1+\alpha})$.
 Moreover, if there exist  some  positive constants $c_i$'s such that
 $d_i^3 p \geq c_i 2^{m^2}$ or $d_i p^3 \leq c_i/2^{m^2}$ for all $i$,
 then \eqref{eq:main1ER} holds with $g(m) = O(c^m)$ for some constant $c>0$.
	\label{thm:main_ER}
\end{thm}
A proof  of Theorem \ref{thm:main_ER} is provided in a longer version of this paper. 
This guarantee shows a similar bias-variance tradeoff, with similar dependence on $m$, which controls the
computational complexity and $\alpha$ which is the normalized spectral gap of the original graph Laplacian.
The main difference in this generic setting is how computational complexity depends on $m$. Since we need to estimate  $\theta_w(G_\Omega)$  which is an unbiased estimate of  $\prod_{u\in w} d_u^{s_u}$, we need to compute it separately for each observed walk $w$ of length $k$ that involves at least one self loop. For the other walks, we exploit
a recent algorithm in counting patterns from \cite{KO17} inspired by
a celebrated result from \cite{AYZ97},
and compute their weighted counts.
This can be made as a look-up table, and overall the complexity scales as
$O(\cc(G)\times \wgt_{\rm max}^3)$ for $m\leq 7$
and for larger $m$ scales as
$O(\cc(G) \times \wgt_{\rm max}^{m/2} \times 2^{m/2})$.
If one has faster algorithms for counting patterns those can be seamlessly included
in the procedure,
for example using recent advances in
recursive methods for counting structures from \cite{curticapean2017homomorphisms}.
Our code is publicly available at {\em url-anonymized}.

\begin{figure}[h]
	\centering
    	\includegraphics[width=0.45\textwidth]{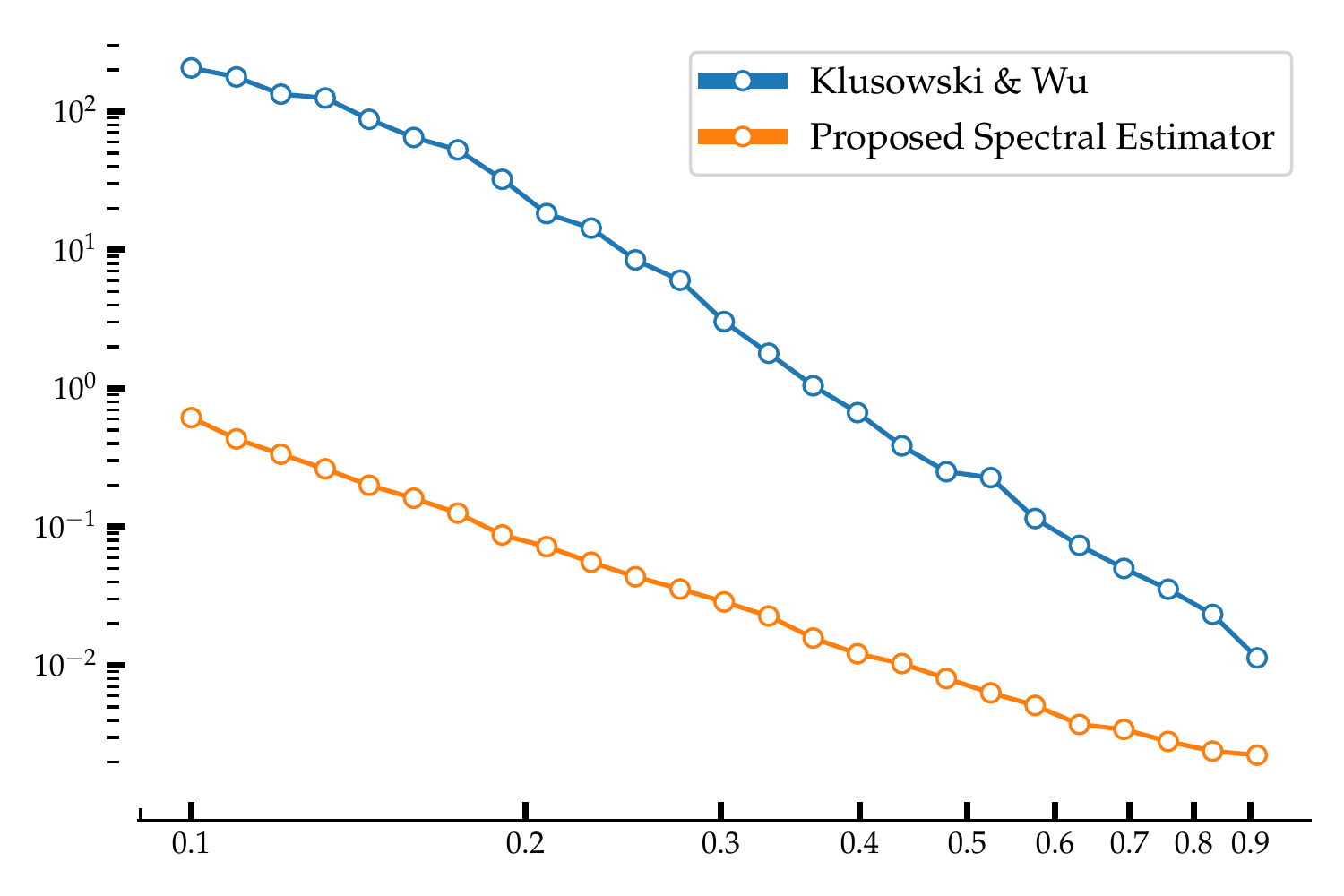}
	\put(-240,35){ \rotatebox{90}{$\frac{\E[ |\widehat{\cc}(G_\Omega,\alpha,\beta,m)-\cc(G)|^2 ]^{1/2}}{|\cc(G)|} $}}
    	\put(-160,-5){vertex sampling probability $p$}
	\caption{The proposed estimator $\hcc(G_\Omega, 0.5 ,80,5)$ improves upon a competing estimator from \cite{KW17}.}
	\label{fig:main_er}
\end{figure}

We run  experiments in Figure \ref{fig:main_er} on
a graph of size $N=5000$ with 50 components
each drawn from Erd\"os-R\'enyi graph with probability half $G_{100,0.5}$.
A moderate $m=5$ is sufficient to achieve multiplicative error
as small as 0.002,
which implies that we make a small mistake
in one out of ten instances.
Note that $\hcc_{\rm chordal}$ and $\hcc_{\rm clique}$ cannot be applied
as the observed subgraph is neither cliques nor chordal. 
A heuristic is proposed in \cite{KW17}, which is explained in Section~\ref{sec:real} 
such that $\hcc_{\rm chordal}$ can be applied. 
As the bias does not depend on $p$,
this experiment implies that with only
$m=5$ the bias is already smaller than $0.002$ and
the variance is dominating. This is due to the fact that union of Erd\"os-R\'enyi
graphs exhibit large spectral gaps.
The variance decreases linearly in this log-log scale,
with respect to the sampling probability $p$.

For the example of union of Erd\"os-R\'enyi graphs
$G_{n,q}$ of the same size with $\cc(G)=N/n$ connected components,
the (normalized) spectral gap is
$\Omega(1)$ for a large enough $q$.
This exhibits the desired spectral gap, as long as $q$ is sufficiently large,
e.g.~$q=\Omega(\log n / n)$.
The ideal case is when $q=1$, which recovers the union of cliques.
The normalized spectral gap is one, which is the maximum possible value.
On the other hand, the spectral gap can be also made arbitrarily small.
Consider a union of $n$-cycles, where each component is a cycle of length $n$.
In this case, the normalized spectral gap scales as $O(1/n^2)$, which can be quite small.
For general graphs,
the difficulty (both in computational complexity and sample complexity)
depends on the spectral gap of the original graph.
If spectral gap is small, then we need higher degree polynomial approximation functions
to make the bias small,
which in turn requires larger patterns to be counted.
This  increases the computational complexity and
also the variance in the estimate.
More samples are required to account for this increased variance.

%
%

\subsection{Real-world graphs}
\label{sec:real}

We run our estimator on two real-world graphs.
\texttt{ACL} \cite{acldata} is an academic citation network that consists of 115,311 citations between 
18,664 papers papers published in ACL  (Association of Computational Linguistics) conferences, 
journals and workshops between years 1965 and 2016. 
\texttt{Hep-PH} \cite{gehrke2003overview} is an academic citation network comprising 412,533 citations 
between 34,546 Hep-PH (high energy physics phenomenology) 
arXiv manuscripts between years 1992 and 2002.

Estimating connected components on real graphs is challenging as they have large condition numbers and 
long cycles. In \cite{KW17}, this is dealt with by triangulating the real graph to turn it into a chordal graph before sampling. 
Alternatively, in this section, to make the estimation problem tractable, we apply the following two modifications to real-world graphs. 
First, we add random edges between nodes that belong to the 
same connected component to improve connectivity.  
If a component has $m$ edges, we add extra $q m$ edges at random. 
We choose $q=0.4$ for \texttt{ACL} and $q=0.6$ for \texttt{Hep-PH}.
Secondly, we trim the degree of extremely high-degree nodes in the network. 
If a node has degree larger than the 95 percentile of the degree distribution, 
we randomly remove its edges until it reaches the degree of the 95 percentile. 
For our experiments, 
we use the first 5000 nodes in each dataset, for computational efficiency. 
After modifying the graphs, the number of connected components in \texttt{ACL} and \texttt{Hep-PH} are 118 and 133.

\begin{figure}[h]
	\centering
    	\includegraphics[width=0.42\textwidth]{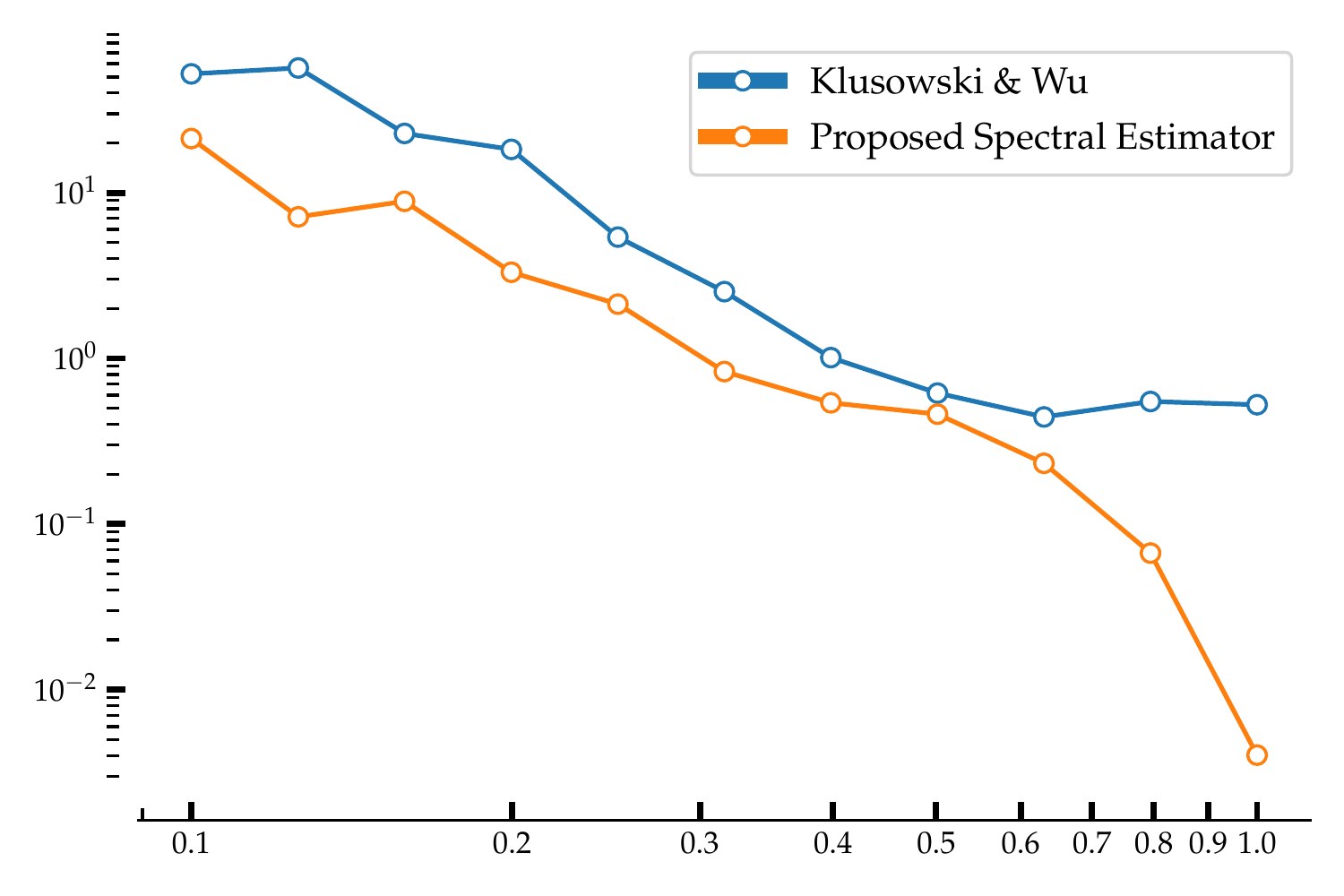}
	\put(-210,15){ \rotatebox{90}{$\frac{\E[ |\widehat{\cc}(G_\Omega,\alpha,\beta,m)-\cc(G)|^2 ]^{1/2}}{|\cc(G)|} $}}
    	\put(-160,-5){vertex sampling probability $p$} 
	\hspace{0.1cm}
	\includegraphics[width=0.42\textwidth]{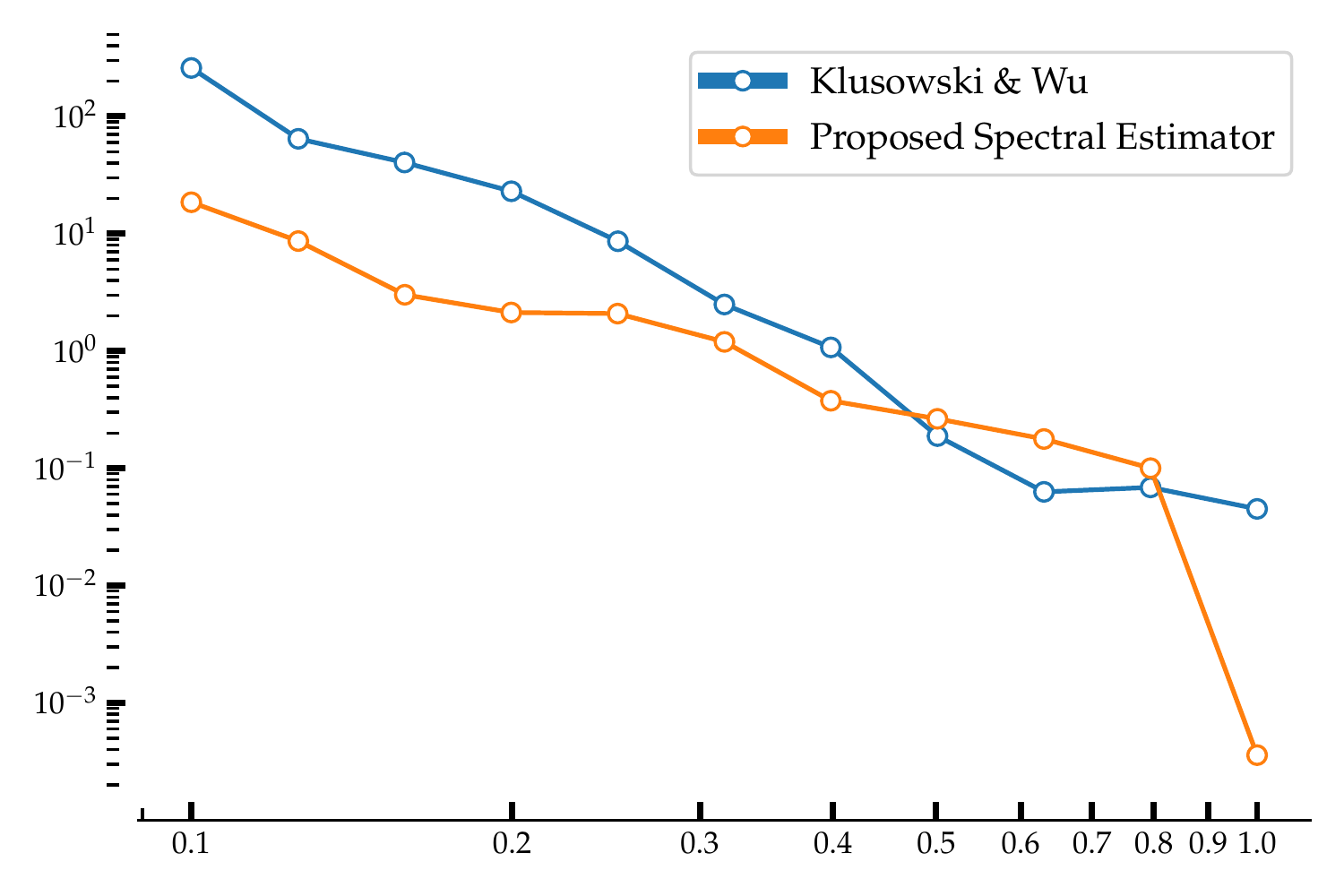}
	\put(-210,15){ \rotatebox{90}{$\frac{\E[ |\widehat{\cc}(G_\Omega,\alpha,\beta,m)-\cc(G)|^2 ]^{1/2}}{|\cc(G)|} $}}
    	\put(-160,-5){vertex sampling probability $p$}
	\caption{On (modified) real-world graphs, \texttt{ACL} (left) and \texttt{Hep-PH} (right) , the proposed approach improves upon the competing baseline approach for 
	most regime of the sampling probability $p$.}
	\label{fig:real}
\end{figure}

We compare the performance of the proposed algorithm 
on two real-world datasets to the algorithm proposed in \cite{KW17}. 
As this estimator is customized for chordal graphs, 
 we first triangulate the subsampled \texttt{ACL} and \texttt{Hep-PH} via a minimal chord completion 
algorithm based on maximum cardinality search \cite{berry2004maximum} and then
apply their smoothed estimator on subsampled graphs of the real-world networks, as proposed in \cite{KW17}. 
Each data point in Figure~\ref{fig:real} is averaged over 50 random instances of subsampling. 
Due to the triangulation, the approach of \cite{KW17} is biased even when the sampling probability is close to one.  
For most of the values of $p$, the proposed estimator outperforms the baseline approach. 
We set $m=5$ for the proposed spectral approach.

\section{Conclusion}

We address the problem of estimating the number of connected components
in an undirected simple graph, when
only a subgraph is observed, where the nodes are chosen uniformly at random.
Existing methods relied on special structures of the graphs,
such as union of disjoint cliques,
union of disjoint trees, and chordal graphs.
Applying a key insight of viewing the number of connected components as
a spectral property that depends on the singular values of the graph Laplacian,
we propose a novel spectral approach to this problem.
Based on the fact that the number of connected components are
the number of zero-valued singular values of the graph Laplacian,
we make several innovations. First,
we propose weighted count of small patterns (which are called network motifs)
to estimate the $k$-th Schatten norm of the graph Laplacian.
Next, to get an estimate of the (monomials of the) degrees,
we propose a novel partitioning scheme that gives an unbiased estimate of the
desired quantity to be used in the estimate of the $k$-th Schatten norms.
We propose a polynomial approximation of the linearly interpolated step function,
and prove a upper bound on the approximation guarantee.
Putting these together, we introduce the first
estimator with provable performance guarantees,
that works for  graphs with positive spectral gaps.

We next discuss several challenges in applying this framework to real world graphs.

\bigskip\noindent
{\bf Counting patterns.}
When the underlying graph $G$ is a disjoint union of cliques, computational complexity of our estimator is $O( m \times \cc(G))$. In this case, for any clique of size $k$, count of all possible patterns in it and the estimates of $\theta_w(G_\Omega)$  for any walk $w$,  characterized by the degrees of the self loops it involves, can be pre-computed and stored in a table for look-up at the time of execution.
$\theta_w(G_\Omega)$  is an unbiased estimate of of the polynomial of the node degrees $\prod_{u\in w} d_u^{s_u}$.
For general graphs, to compute $\hTheta_k(G_{\Omega}) $, we need to compute $\theta_w(G_\Omega)$  separately for each observed walk $w$ that has at least one self loop. For the walks that do not involve any self-loop, we can use matrix multiplication based pattern counting algorithms proposed in \cite{KO17} for $m \leq 7$.  For $m > 7$, one can use homomorphism based a recent recursive algorithm from \cite{curticapean2017homomorphisms}.
Therefore, for general graphs the major computational complexity arises in computing $\theta_w(G_\Omega)$ for walks $w$ that has at least one self-loop.
In a different sampling scenario, if we have the additional information of the degree of each node that we observe then computing $\theta_w(G_\Omega)$ can be made fast for all the walks $w$.
Another option is to apply recent advances in
sampling-based methods for counting patterns, including
wedge sampling \cite{SPK13},
the 3-path sampling \cite{JSP15}, Moss \cite{WTZ15}, GRAFT \cite{RBA14}, and using Hamiltoniam paths for debiasing \cite{CL16}.
However, it is not immediate how to include the estimation of the monomials of the degrees into these existing fast methods.

\bigskip\noindent
{\bf Other sampling techniques.}
In practical settings, sampling nodes uniformly at random might be unrealistic.
Our estimator  generalizes naturally to a broader class of sampling schemes, which we call
{\em graph sampling}.
Consider a scenario where you first sample an {\em unlabelled} mother graph $H_0$ of the same size as $G$
(the graph of interest)
from any distribution (in particular we do not require any independence on the sampled edges).
Then, we apply a permutation drawn uniformly at random to assign node labels to the unlabelled graph $H_0$.
Let $H$ denote this labelled graph, which we use to sample the original graph of interest $G$.
Specifically, for all edges in $H$, we observe whether the corresponding edge is present or not in $G$.
Namely, we observe the adjacency matrix of $G$, but masked by the adjacency matrix of $H$.
The random permutation ensures that the sampling probability for a pattern only depends on the shape of the pattern
and not the specific labels of the nodes involved,
making our algorithm extendible up to properly applying the debiasing as per the new sampling model.
The model studied in this paper
is a special case of graph sampling where $H_0$ is a clique of random size, drawn according to a binomial distribution.

On the other hand, more practical sampling scenarios are adaptive to the topology of the graph, creating selection biases.
Examples include crawling a connected path from a starting node, sampling higher degree nodes, or
sampling via random walks. These create dependencies among the topology and the sampling, which we believe is outside the scope of this paper, but nevertheless poses an interesting new research direction.



 \section{Proofs}

 We provide the proof sketch of the  main results.
Recall
$
	\hcc(G_\Omega,\alpha,\beta,m) =  N   - \sum_{k = 1}^m \frac{a_k}{\beta^k} \hTheta_k(G_\Omega)
$,
where $\hTheta_k(G_\Omega)$ is an unbiased estimate of Schatten-$k$ norm of $L$ defined in \eqref{eq:estimate} and $a_k$'s are coefficients of polynomial $f_{b}(x)$ defined in \eqref{eq:defpoly}. We show in \eqref{eq:prf2} that for the proposed estimator, bias is bounded as
\begin{align} \label{eq:bias}
 \big|\cc(G) - \E\big[\hcc(G_\Omega,\alpha,\beta,m)\big]\big|  &\leq
 (N - \cc(G)) ((1-\alpha)/(1+\alpha) )^\mmm \nonumber \\
& = (N - \cc(G))\gamma^m\,,
\end{align}
where $\gamma \equiv (1-\alpha)/(1+\alpha)$.
For the choice of $\tb\equiv (2/(1+\alpha))[1,\ldots,1]$, the coefficient of $x^k$ in $f_b(x)$ is bounded as $|a_k| \leq {m \choose k}2^k \leq 2^m 2^k\leq 4^m$. Therefore the mean square error is bounded by
\begin{align} \label{eq:mse}
&\E\big[ ( \widehat{\cc}(G_\Omega,\alpha,\beta,m) - \cc(G) )^2 \big] \; = \nonumber \\
& \; 2^{4m}{\rm Var}\bigg(\sum_{k=1}^m \frac{1}{\beta^k} \hTheta_k(G_\Omega) \bigg) + \Big((N - \cc(G))\gamma^m\Big)^2\,.
\end{align}

Since the Schatten norm estimator is unbiased, $\E[\hTheta_k(G_\Omega)] = \norm{L}_k^k$, we have
$\E\big[\hcc(G_\Omega,\alpha,\beta,m)\big] 
 =  N - \sum_{i = 1}^N f_{\alpha}(\sigma_i(\L))
$.
where $f_{\alpha}$ is defined in Equation \eqref{eq:genrankapproximation}. Note that $\beta$ is chosen such that the non-zero eigenvalues of $\widetilde L = L/\beta$ are bounded between $\alpha$ and 1. With the proposed choice of polynomial function $f_{\alpha}$, we have $f_{\alpha} = f_{b}$. Using, Equation \eqref{eq:eq3},
$\cc(G) = N - \sum_{i = 1}^N H_{\alpha}( \sigma_i(\L))$,
along with $\max_{x\in[\alpha,1]} |H_\alpha(x) - f_{\tb}(x)|  = ( (1-\alpha)/(1+\alpha) )^\mmm $, where $\tb\equiv (2/(1+\alpha))[1,\ldots,1]$, we have,
\begin{align}
&\Big|\cc(G) - \E\big[\hcc(G_\Omega,\alpha,\beta,m)\big]\Big|  =  \sum_{i=1}^N \Big( H_{\alpha}( \sigma_i(\L)) - f_{\tb}(x) \Big) \nonumber \\
& =  \sum_{i  : \sigma_i(\L) = 0 } \Big( H_{\alpha}( \sigma_i(\L)) - f_{\tb}(x) \Big) + \sum_{i  : \sigma_i(\L) \neq 0 } \Big( H_{\alpha}( \sigma_i(\L)) - f_{\tb}(x) \Big) \nonumber \\
& \leq  \cc(G) \Big( H_{\alpha}(0) - f_{\tb}(0) \Big) + (N - \cc(G))\max_{x\in[\alpha,1]} |H_\alpha(x) - f_{\tb}(x)| \nonumber\\
& \leq   (N - \cc(G)) ((1-\alpha)/(1+\alpha) )^\mmm = (N - \cc(G))\gamma^m\,.\label{eq:prf2}
\end{align}

For the two cases: $(a)$ when the underlying graph $G$ is disjoint union of cliques, and $(b)$ a general graph $G$ with maximum degree $\dmax$, we provide bounds on the variance of the Schatten $k$-norm estimator that leads to the bounds on mean square error using Equation \eqref{eq:mse}.

Denote each connected component of $G$ by $G^{(i)}$ for $1 \leq i \leq \cc(G)$, and let $G_{\Omega}^{(i)}$ denote the randomly observed subgraph of the connected component $G^{(i)}$. Then, we have,
\begin{eqnarray}\label{eq:break_var}
{\rm Var}\bigg(\sum_{k=1}^m \frac{1}{\beta^k} \hTheta_k(G_\Omega) \bigg) = \sum_{i=1}^{\cc(G)} {\rm Var}\bigg(\sum_{k=1}^m \frac{1}{\beta^k} \hTheta_k(G_\Omega^{(i)}) \bigg)\,.
\end{eqnarray}
Note that, our estimator of Schatten $k$-norm  naturally decomposes, and can be computed separately for each connected component $G^{(i)}$ and then added together to get the estimate for the graph $G$.

\subsection{Proof of Theorem \ref{thm:main}}
\label{sec:main_proof}
The following lemma provides bound on the  variance of  Schatten $k$-norm estimator  for a clique graph. We give a proof in Section \ref{sec:clique}.

\begin{lemma} For a clique graph $G$ on $\wgt$ vertices, there exists a universal positive constant $C$ such that for $\wgt \geq C$, variance of Schatten $k$-norm estimator $\hTheta_k(G_\Omega)$ is bounded by
\begin{eqnarray} \label{eq:lem_clique1}
{\rm Var} \big(\hTheta_k(G_\Omega)\big) & \leq & g(k) \frac{\wgt^{2k+2}}{p}\Big( 1 + \frac{1}{(\wgt p)^{2k-1}}\Big)\,,
\end{eqnarray}
where $g(k) = O(k!)$. Moreover, if there exists a positive constant $c$ such that $\wgt^3 p \geq c k!$ or $\wgt p^3 \leq c/k!$ then \eqref{eq:lem_clique1} holds with $g(k) = {\rm poly}(k)$. \label{lem:clique}
\end{lemma}
Using Equations \eqref{eq:mse} and \eqref{eq:break_var} along with Lemma \ref{lem:clique}, and the fact that $	\beta \geq \wgt$, Theorem \ref{thm:main} follows immediately.
For a proof of this lemma, we refer to the longer version of this paper. 



 \bibliographystyle{plain}
\bibliography{cc}

\begin{thebibliography}{10}

\bibitem{AYZ97}
N.~Alon, R.~Yuster, and U.~Zwick.
\newblock Finding and counting given length cycles.
\newblock {\em Algorithmica}, 17(3):209--223, 1997.

\bibitem{BKM14}
P.~Berenbrink, B.~Krayenhoff, and F.~Mallmann-Trenn.
\newblock Estimating the number of connected components in sublinear time.
\newblock {\em Information Processing Letters}, 114(11):639--642, 2014.

\bibitem{berend2010improved}
Daniel Berend and Tamir Tassa.
\newblock Improved bounds on bell numbers and on moments of sums of random
  variables.
\newblock {\em Probability and Mathematical Statistics}, 30(2):185--205, 2010.

\bibitem{berry2004maximum}
Anne Berry, Jean~RS Blair, Pinar Heggernes, and Barry~W Peyton.
\newblock Maximum cardinality search for computing minimal triangulations of
  graphs.
\newblock {\em Algorithmica}, 39(4):287--298, 2004.

\bibitem{CRT05}
B.~Chazelle, R.~Rubinfeld, and L.~Trevisan.
\newblock Approximating the minimum spanning tree weight in sublinear time.
\newblock {\em SIAM Journal on computing}, 34(6):1370--1379, 2005.

\bibitem{Che1853}
Pafnuti~Lvovich Chebyshev.
\newblock {\em Th{\'e}orie of the mechanisms known as parallel {\ e}
  logrammes}.
\newblock Critical Academy of Science Printing, 1853.

\bibitem{CL16}
Xiaowei Chen and John~CS Lui.
\newblock Mining graphlet counts in online social networks.
\newblock In {\em Data Mining (ICDM), 2016 IEEE 16th International Conference
  on}, pages 71--80. IEEE, 2016.

\bibitem{curticapean2017homomorphisms}
Radu Curticapean, Holger Dell, and D{\'a}niel Marx.
\newblock Homomorphisms are a good basis for counting small subgraphs.
\newblock {\em arXiv preprint arXiv:1705.01595}, 2017.

\bibitem{Fra78}
O.~Frank.
\newblock Estimation of the number of connected components in a graph by using
  a sampled subgraph.
\newblock {\em Scandinavian Journal of Statistics}, pages 177--188, 1978.

\bibitem{Fra82}
O.~Frank and F.~Harary.
\newblock Cluster inference by using transitivity indices in empirical graphs.
\newblock {\em Journal of the American Statistical Association},
  77(380):835--840, 1982.

\bibitem{gehrke2003overview}
Johannes Gehrke, Paul Ginsparg, and Jon Kleinberg.
\newblock Overview of the 2003 kdd cup.
\newblock {\em ACM SIGKDD Explorations Newsletter}, 5(2):149--151, 2003.

\bibitem{Goo49}
L.~A. Goodman.
\newblock On the estimation of the number of classes in a population.
\newblock {\em The Annals of Mathematical Statistics}, pages 572--579, 1949.

\bibitem{JSP15}
Madhav Jha, C~Seshadhri, and Ali Pinar.
\newblock Path sampling: A fast and provable method for estimating 4-vertex
  subgraph counts.
\newblock In {\em Proceedings of the 24th International Conference on World
  Wide Web}, pages 495--505. International World Wide Web Conferences Steering
  Committee, 2015.

\bibitem{KO17}
A.~Khetan and S.~Oh.
\newblock Spectrum estimation from a few entries.
\newblock {\em arXiv preprint arXiv:1703.06327}, 2017.

\bibitem{KW17}
J.~M. Klusowski and Y.~Wu.
\newblock Estimating the number of connected components in a graph via subgraph
  sampling.
\newblock {\em Technical report}, 2017.
\newblock available at http://www.stat.yale.edu/$\sim$yw562/preprints/cc.pdf.

\bibitem{MH02}
John~C Mason and David~C Handscomb.
\newblock {\em Chebyshev polynomials}.
\newblock CRC Press, 2002.

\bibitem{PM72}
T~Parks and J~McClellan.
\newblock Chebyshev approximation for nonrecursive digital filters with linear
  phase.
\newblock {\em IEEE Transactions on Circuit Theory}, 19(2):189--194, 1972.

\bibitem{acldata}
Dragomir~R. Radev, Pradeep Muthukrishnan, Vahed Qazvinian, and Amjad Abu-Jbara.
\newblock The acl anthology network corpus.
\newblock {\em Language Resources and Evaluation}, pages 1--26, 2013.

\bibitem{Raf05}
D.~Rafiei.
\newblock Effectively visualizing large networks through sampling.
\newblock In {\em Visualization, 2005. VIS 05. IEEE}, pages 375--382. IEEE,
  2005.

\bibitem{RBA14}
Mahmudur Rahman, Mansurul~Alam Bhuiyan, and Mohammad Al~Hasan.
\newblock Graft: An efficient graphlet counting method for large graph
  analysis.
\newblock {\em IEEE Transactions on Knowledge and Data Engineering},
  26(10):2466--2478, 2014.

\bibitem{SPK13}
Comandur Seshadhri, Ali Pinar, and Tamara~G Kolda.
\newblock Triadic measures on graphs: The power of wedge sampling.
\newblock In {\em Proceedings of the 2013 SIAM International Conference on Data
  Mining}, pages 10--18. SIAM, 2013.

\bibitem{WTZ15}
Pinghui Wang, Jing Tao, Junzhou Zhao, and Xiaohong Guan.
\newblock Moss: A scalable tool for efficiently sampling and counting 4-and
  5-node graphlets.
\newblock {\em arXiv preprint arXiv:1509.08089}, 2015.

\bibitem{ZWJ15}
Y.~Zhang, M.~J. Wainwright, and M.~I. Jordan.
\newblock Distributed estimation of generalized matrix rank: Efficient
  algorithms and lower bounds.
\newblock {\em arXiv preprint arXiv:1502.01403}, 2015.

\end{thebibliography}

\newpage
\appendix
\section*{Appendix}
\section{Proofs}


\subsection{Proof of Lemma \ref{lem:clique}}
\label{sec:clique}

\begin{figure}[h]
	\begin{center}
	\includegraphics[width=.4\textwidth]{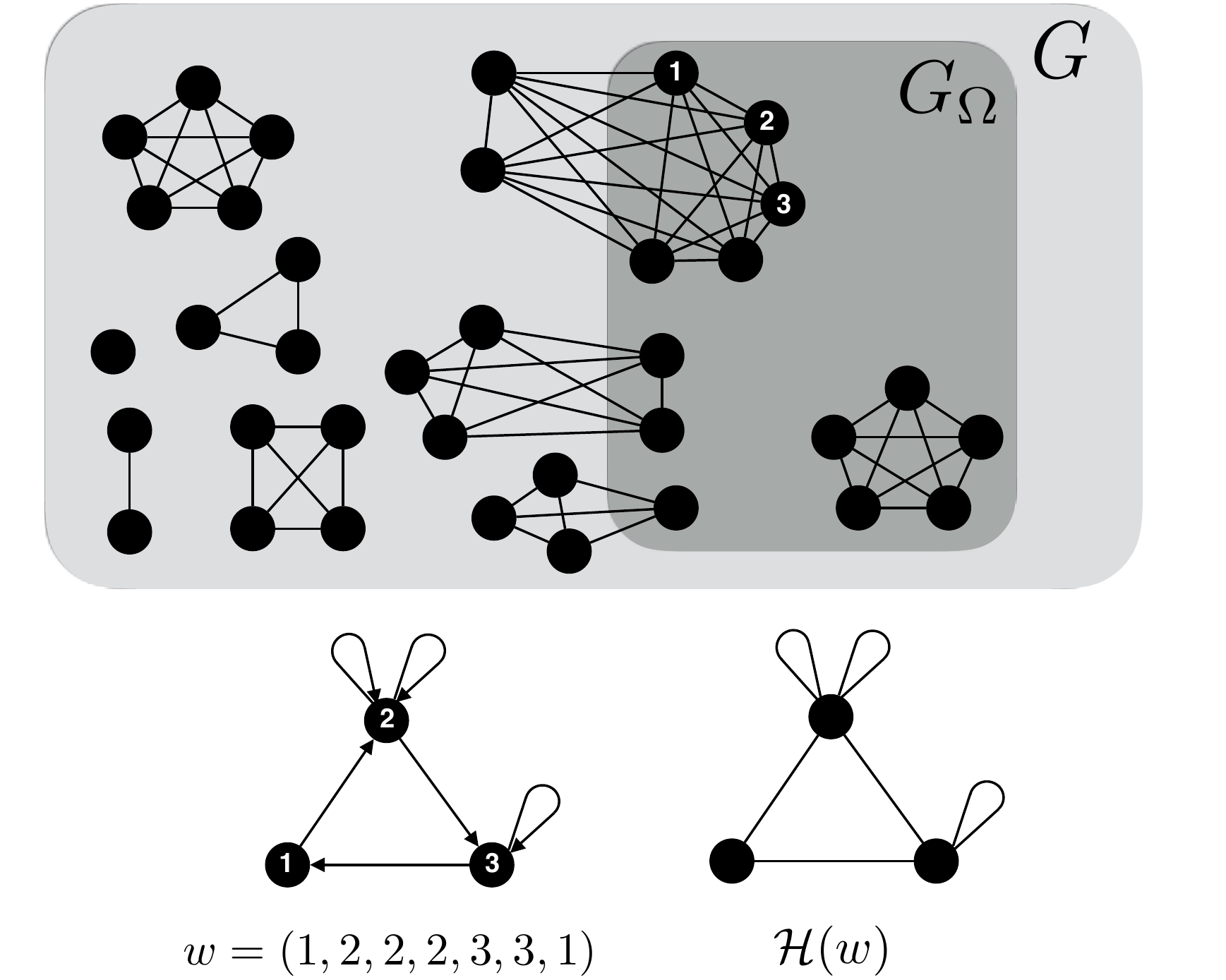}
	\end{center}
	\caption{For a set of cliques $G$,
	we get a sampled  $G_\Omega$.
	An example of a length-$(k=6)$ closed walk $w=(1,2,2,2,3,3,1)$
	and its corresponding $k$-cyclic pseudograph $\cH(w) \in \bH_6$.}
	\label{fig:clique}
\end{figure}

For a $k$-cyclic pseudograph $H = (V_H,E_H)$, let $S_H$ denote the set of  self-loops in $H$.
Recall that in \eqref{eq:estimate} Schatten $k$-norm estimator of Laplacian $L$ is given by
\begin{eqnarray}
	   \hTheta_k(G_{\Omega})  =
	   \sum_{H \in \bH_k} \frac{(-1)^{k-|S_H|} }{p^{|V_H|}}  \Big\{ \sum_{w: \cH(w)=H} \theta(w,G_\Omega){\mathbb I}(w\subseteq G_\Omega)  \Big\} \;. \nonumber
\end{eqnarray}

For a clique graph $G$, the analysis of the above estimator simplifies significantly.
We illustrate this with an example in Figure \ref{fig:clique}.
Consider a length $k=6$ walk $w=(1,2,2,2,3,3,1)$ with a corresponding $k$-cyclic pseudograph $\cH(w)$.
In general, the degree estimator $\theta(w,G_\Omega)$ is chosen such that
$\E[\theta(w,G_\Omega)] = d_2^2 d_3$ where $d_2$ and $d_3$ are the degrees of nodes $2$ and $3$ in $G$, respectively.
This simplifies significantly for a clique graph due to the fact that
the degree of those nodes in a closed walk $w$ are the same. Note that our estimator is general, and does not use
this information or the fact that the underlying graph component is a clique. It is only the analysis that simplifies.
Therefore, for a clique graph $G$, the degree estimator $\theta(w,G_\Omega)$ satisfies
$\E[\theta(w,G_\Omega) | w \subseteq G_{\Omega}] = (\wgt-1)^{|S_H|}$, where $\wgt$ is the size of the clique $G$.
In the example, we have $\wgt=7$ and $|S_H|=3$, therefore  $\E[\theta(w,G_\Omega)]=6^3$.
Hence, it is best to further partition $\bH_k$ according to
the number of nodes $\ell=|V_H|$ and the number of self-loops $s=|S_H|$.
Precisely, we define
	\begin{eqnarray*}
		\bH_{k,\ell,s} \;\; \equiv \;\; \{ H(V_H,E_H) \in \bH_k \,:\, |V_H| = \ell \text{ and } |S_H|=s \} \;,
	\end{eqnarray*}
for $\ell=1$, $s = k$ and $2 \leq \ell \leq k$, $0 \leq  s \leq k-\ell$.
There are total
$|\{w\in W\,:\, \cH(w)\in \bH_{k,\ell,s}\}| \leq  {\ell^{2k-s}} \wgt^{\ell}$ corresponding walks in this set.
Here,  $W$ denotes the collection of all length $k$ closed walks on a complete graph of $\wgt$ vertices.
We slightly overload the notion of complete graph to refer to an undirected graph with
not only all the ${\wgt(\wgt-1)/2}$ simple edges but also with $\wgt$ self loops as well.
when $G$ is a clique graph, the estimator \eqref{eq:estimate} can be re-written as
\begin{align}
	\label{eq:pf1}
	   & \hTheta_k(G_{\Omega}) \;\; =\;\; \sum_{\ell = 1}^k  \sum_{s = 0}^{k-\ell+1} \sum_{H \in \bH_{k,\ell,s}}
	   \Big\{ \frac{(-1)^{k-s}}{p^{\ell}} \, \times \nonumber\\
	   & \;\;\;\;\;\;\;\;\;\;\;\;\;\;\;\;  \sum_{w \in W: \cH(w)=H}   \,  \theta(w,G_{\Omega}) {\mathbb I}(w\subseteq G_\Omega) \Big\} \,.
\end{align}

Given this unbiased estimator, we provide an upper bound on the variance of each of the partitions.
For any two walks $w, w' \in  W$, let $|w \cap w'|$ denote the number of overlapping unique vertices of walks $w$ and $w'$. We have,
\begin{align}
&   {\rm Var}\big(\hTheta_k(G_\Omega))
 \; =\;    \sum_{\ell = 1}^k  \sum_{s = 0}^{k-\ell+1} \sum_{H \in \bH_{k,\ell,s}} \Big\{  \frac{1}{p^{2\ell}}  \nonumber\\
 & \hspace{3cm} \times\sum_{w \in W: \cH(w)=H}  {\rm Var} \Big(  \theta(w,G_{\Omega}) {\mathbb I}(w\subseteq G_\Omega)   \Big) \Big\} \nonumber\\
 &\;\;\; + \;\;2\sum_{\tilde \ell=0}^{k}\sum_{\substack{w, w' \in W\\ |w \cap w'| = \tilde \ell}}
	   {\rm Cov} \Big(  \theta(w,G_{\Omega})  \mathbb{I}(w \subseteq G_\Omega)\,,   \theta(w',G_{\Omega}) \mathbb{I}(w' \subseteq G_\Omega)  \Big)   \nonumber\\
	    & \hspace{3cm} \times\frac{(-1)^{|S_{\cH(w)}| + |S_{\cH(w')}|}}{p^{|V_{\cH(w)}|+ |V_{\cH(w')}|}}\,.\label{eq:pf2}
\end{align}

The following technical lemma provides upper bounds on the variance and covariance terms.
We provide a proof in Section \ref{sec:proof_clique_var_0}.

\begin{lemma}
	\label{lem:clique_var_0}
Under the hypothesis of Lemma \ref{lem:clique}, for a length-$k$ walk $w$ over $\ell$ distinct nodes with $s \geq 1$ self-loops, the following holds:
	\begin{eqnarray}
{\rm Var} \Big(  \theta (w,G_{\Omega}) {\mathbb I}(w\subseteq G_\Omega)   \Big)
 \leq  f(\ell,s) \Big( p^{\ell-1}{\wgt}^{2s-1} + \wgt p^{\ell+1-2s} \Big)   + p^{\ell}{\wgt}^{2s} ,
 \label{eq:clique_var}
\end{eqnarray}
and when $\ell =1$, $s = k$, we have,
	\begin{eqnarray}
{\rm Var} \Big(  \theta (w,G_{\Omega}) {\mathbb I}(w\subseteq G_\Omega)   \Big)
 \leq  f(k) {\wgt}^{2k-1} + g(k)\wgt p^{2-2k}  + p{\wgt}^{2k} , \label{eq:pf8}
\end{eqnarray}
and for any length-$k$ walks $w_1, w_2$ over $\ell_1, \ell_2$ distinct nodes  with $\tilde\ell$ unique overlapping nodes, $|w_1 \cap w_2| = \tilde\ell $, $s_1,s_2 \geq 1$ self-loops respectively, the covariance term can be upper bounded by:
\begin{eqnarray}
	{\rm Cov} \Big(  \theta(w_1,G_{\Omega})  \mathbb{I}(w_1 \subseteq G_\Omega)\,,   \theta(w_2,G_{\Omega}) \mathbb{I}(w_2 \subseteq G_\Omega)  \Big)  \nonumber \\
	\leq  f(\ell',s')p^{((\ell_1+\ell_2 - \tilde\ell)-(s_1+s_2))}\Big((\wgt p)^{(s_1+s_2)} + \wgt p \Big)\,, \label{eq:pf7}
\end{eqnarray}
for some function $f(\ell,s)=O(k!)$, $g(k) = {\rm poly}(k)$, where $p$ is the vertex sampling probabiliy. $\ell' \equiv \max\{\ell_1,\ell_2\}$ and $s' \equiv \max\{s_1,s_2\}$.
\end{lemma}

We use this lemma to get bound on ${\rm Var}\big(\hTheta_k(G_\Omega))$. First, we get a bound on the total variance term.
For a walk $w \in W$ with $\cH(w)\in \bH_{k,\ell,s}$ with $1 \leq s \leq k-2$, using \eqref{eq:clique_var}, we have,
\begin{align}
& \frac{\wgt^{\ell}}{p^{2\ell}}{\rm Var} \Big(  \theta(w,G_{\Omega}) {\mathbb I}(w\subseteq G_\Omega)\Big) \nonumber \\
& \hspace{1cm}\leq  f(\ell,s) \Big( \frac{{\wgt}^{\ell+2s-1}}{p^{\ell+1}} + \frac{\wgt^{\ell+1}}{p^{\ell + 2s -1}} \Big)   \;+\; \frac{{\wgt}^{2s+\ell}}{p^{\ell}}\nonumber\\
& \hspace{1cm}\leq  f(k) \Big(\frac{\wgt^{2k-3}}{p^3} + \frac{\wgt^{3}}{p^{2k -3}} \Big) \;+ \; \frac{\wgt^{2k-2}}{p^2}  \,. \label{eq:pf9}
\end{align}

For a walk $w \in W$ with $\cH(w)\in \bH_{k,\ell,s}$ with $\ell=1$, $s = k$, using \eqref{eq:pf8}, we have,
\begin{align}
&\frac{\wgt^{\ell}}{p^{2\ell}}{\rm Var}\Big(  \theta(w,G_{\Omega}) {\mathbb I}(w\subseteq G_\Omega)\Big) \nonumber \\
& \hspace{1cm}\leq  f(k) {\wgt}^{2k}p^{-2} \;+\; g(k)\wgt^2 p^{-2k} \;+ \; w^{2k+1}p^{-1} \,. \label{eq:pf10}
\end{align}

For a walk $w$ with $s=0$, $ \theta(w,G_{\Omega}) = 1$, and, we have,
\begin{eqnarray}
\frac{\wgt^{\ell}}{p^{2\ell}}{\rm Var}\Big(  \theta(w,G_{\Omega}) {\mathbb I}(w\subseteq G_\Omega)\Big) & \leq & \frac{\wgt^\ell}{p^{\ell}}\,. \label{eq:pf11}
\end{eqnarray}

Combining, Equations \eqref{eq:pf9}, \eqref{eq:pf10}, and \eqref{eq:pf11}, and using
$|\{w\in W\,:\, \cH(w)\in \bH_{k,\ell,s}\}| \leq {\ell^{2k-s}} {\wgt}^{\ell}$, we have
\begin{align}
&\sum_{\ell = 1}^k  \sum_{s = 0}^{k-\ell+1} \sum_{H \in \bH_{k,\ell,s}} \Big\{  \frac{1}{p^{2\ell}} \, \sum_{w \in W: \cH(w)=H} \mathbb{I}(w \subseteq G)\, \times \nonumber\\
&\hspace{3cm}{\rm Var} \Big(  \theta(w,G_{\Omega}) {\mathbb I}(w\subseteq G_\Omega)   \Big) \Big\} \nonumber\\
& \hspace{1cm}\leq  f(k)\wgt^2 p^{-2k} \;+\; \wgt^{2k+1} p^{-1}\,,\label{eq:pf105}
\end{align}
and if $\wgt p^3 \leq 1/f(k)$, then the above quantity is bounded by $g(k)\wgt^2p^{-2k}(1 + o(1))$.
If $\wgt^3 p \geq f(k)$, then the above quantity is bounded by $\wgt^{2k+1}p^{-1}(1 + o(1))$.

Consider covariance term of two length-$k$ walks $w_1, w_2$ over $\ell_1, \ell_2$ distinct nodes  with $\tilde\ell$ unique overlapping nodes, $|w_1 \cap w_2| = \tilde\ell $, and $s_1,s_2 \geq 1$ self-loops with $s_1 + s_2 < 2k$. Since there are a total of $f(k)\wgt^{\ell_1 + \ell_2 -\tilde\ell}$ such walks, using \eqref{eq:pf7}, we have
\begin{align}
&\wgt^{\ell_1 + \ell_2 - \tilde\ell} 	   {\rm Cov} \Big(  \theta(w_1,G_{\Omega})  \mathbb{I}(w_1 \subseteq G_\Omega)\,,   \theta(w_2,G_{\Omega}) \mathbb{I}(w_2 \subseteq G_\Omega)  \Big) \nonumber\\
	    &\;\;\;\;\times \frac{(-1)^{|S_{\cH(w_1)}| + |S_{\cH(w_2)}|}}{p^{|V_{\cH(w_1)}|+ |V_{\cH(w_2)}|}} \label{eq:pf103}\\
	    &\;\;\leq \frac{\wgt^{\ell_1 + \ell_2 - \tilde\ell}}{p^{\tilde\ell + s_1 + s_2}}\bigg( (\wgt p)^{s_1+s_2} + \wgt p\bigg)\label{eq:pf102}\\
	    & \;\;\leq \frac{\wgt^{\ell_1 + \ell_2 + s_1 + s_2}}{(\wgt p)^{\tilde\ell}} + \wgt p\frac{\wgt^{\ell_1+ \ell_2 -\tilde\ell}}{p^{s_1 + s_2 + \tilde\ell}}\label{eq:pf101}\\
	    & \;\;\leq  \wgt^{2k+1} + \wgt p\frac{\wgt^2}{p^{2k-1}}\, \label{eq:pf106}
\end{align}
where in \eqref{eq:pf101}, for the first term the maximum is achieved at $\ell_1= 1,s=k,\ell_2 = 2, s_2 = k-2,\tilde\ell =0$, for the second term the maximum is achieved at same $\ell_1,s_1,\ell_2,s_2$ with $\tilde\ell = 1$. Note that in the expression \eqref{eq:pf102}, the first term is dominating when $\wgt p \geq 1$, and the second term is dominating when $\wgt p <1$.

When $s_1 + s_2 = 2k$ the two walks are self loop walks on single node with $s_1 = s_2 = k$. Since the graph $G$ is a clique graph, $\theta(w,G_{\Omega})$ for self loop walks depends only upon observed size of the clique, and $\theta(w_1,G_{\Omega}) = \theta(w_2,G_{\Omega})$. Therefore, the expression in \eqref{eq:pf103} can be bounded as:
\begin{align}
&\wgt^{\ell_1 + \ell_2 - \tilde\ell} 	   {\rm Cov} \Big(  \theta(w_1,G_{\Omega})  \mathbb{I}(w_1 \subseteq G_\Omega)\,,   \theta(w_2,G_{\Omega}) \mathbb{I}(w_2 \subseteq G_\Omega)  \Big) \nonumber\\
&\;\;\;\;\;\;\;\;
	    \times \frac{(-1)^{|S_{\cH(w_1)}| + |S_{\cH(w_2)}|}}{p^{|V_{\cH(w_1)}|+ |V_{\cH(w_2)}|}} \nonumber\\
&\;\;\;\;\leq  \frac{\wgt^{2\ell}}{p^{2\ell}}{\rm Var}\Big(  \theta(w,G_{\Omega}) {\mathbb I}(w\subseteq G_\Omega)\Big)\nonumber\\
 & \;\;\;\;\leq  f(k) {\wgt}^{2k+1}p^{-2} \;+\; g(k)\wgt^3 p^{-2k} \;+ \; w^{2k+2}p^{-1}\,, \label{eq:pf104}
\end{align}
where \eqref{eq:pf104} follows from \eqref{eq:pf10}.

Lemma \ref{lem:clique} follows immediately by combining \eqref{eq:pf2} with \eqref{eq:pf105}, \eqref{eq:pf106} and \eqref{eq:pf104}.


\subsection{Proof of Lemma \ref{lem:clique_var_0}}
\label{sec:proof_clique_var_0}
We use the following technical lemma to get bounds on conditional variance and covariance of the estimator $\theta$. We provide a proof in Section \ref{sec:proof_clique_var}.
\begin{lemma}
	\label{lem:clique_var}
	Under the hypothesis of Lemma \ref{lem:clique}, for length-$k$ walks $w_1, w_2$ over $\ell_1, \ell_2$ distinct nodes with $s_1,s_2 \geq 1$ self-loops respectively,
	the conditional variance of estimator $\theta(w_1,G_\Omega)$, defined in  \eqref{eq:estimate_theta}, given that all the nodes in the walk are sampled
	can be upper bounded by
	\begin{align}
		& {\rm Var}\Big( \theta(w_1,G_{\Omega})\Big| w_1\subseteq G_\Omega   \Big) \nonumber \\
		 &\;\; \leq \;\;
		  f(\ell_1,s_1) \Big( p^{-1}(\wgt-\ell_1)^{2s_1-1} + (\wgt-\ell_1)p^{1-2s_1} \Big) \;, \text{ and } \label{eq:clique_var22}\\
		  &\E\Big[ \theta(w_1,G_{\Omega})  \theta(w_2,G_{\Omega}) \Big|  \mathbb{I}(w_1 \subseteq G_\Omega)  \mathbb{I}(w_2 \subseteq G_\Omega) \Big] \nonumber \\
		 & \;\; \leq  \;\;
		f(\ell',s')p^{-(s_1+s_2)}\Big((\wgt p)^{s_1+s_2} + \wgt p \Big) \;,
 	\label{eq:clique_cov}
 	\end{align}
		for some function $f(\ell,s)=O(k!)$,
		where $p$ is the vertex sampling probability. $\ell' \equiv \max\{\ell_1,\ell_2\}$ and $s' \equiv \max\{s_1,s_2\}$. Moreover, for a length $k$ walk $w$ with $\ell = 1$, and $s= k$,
			\begin{eqnarray}
		 {\rm Var}\Big( \theta(w,G_{\Omega})\Big| w\subseteq G_\Omega   \Big)
		 		  \leq   \frac{f(k)(\wgt-1)^{2k-1}}{p} + \frac{g(k) (\wgt-1)}{p^{2k-1}} ,  \label{eq:clique_var_selfloops}
    \end{eqnarray}
    for some function $g(k) = {\rm poly}(k)$.
\end{lemma}

Recall for a clique graph, we have $\E[\theta(w,G_\Omega) | w \subseteq G_{\Omega}] = (\wgt-1)^{s}$. Therefore, we have,
\begin{align}
&{\rm Var} \Big(  \theta (w,G_{\Omega}) {\mathbb I}(w\subseteq G_\Omega)   \Big) \nonumber\\
& =  \E\Big[{\rm Var}\Big( \theta(w,G_{\Omega})  {\mathbb I}(w\subseteq G_\Omega)  \Big|  {\mathbb I}(w\subseteq G_\Omega) \Big)  \Big] \nonumber\\
 & \;\;\;\; + \;\; {\rm Var}\Big(\E\Big[\theta(w,G_{\Omega})  {\mathbb I}(w\subseteq G_\Omega)  \Big|  {\mathbb I}(w\subseteq G_\Omega)  \Big] \Big)\nonumber\\
& =  p^{\ell}\; {\rm Var}\Big( \theta (w,G_{\Omega})\Big| w\subseteq G_\Omega   \Big)  \;+\; p^{\ell}(1 - p^{\ell})(\wgt-1)^{2s} \nonumber\\
& \leq  f(\ell,s) \Big( p^{\ell-1}(\wgt-\ell)^{2s-1} + (\wgt-\ell)p^{\ell+1-2s} \Big)   \nonumber\\
& \;\;\;\;+\;\; p^{\ell}(1 - p^{\ell})(\wgt-1)^{2s}\nonumber\\
& \leq  f(\ell,s) \Big( p^{\ell-1}{\wgt}^{2s-1} + \wgt p^{\ell+1-2s} \Big)   \;+\; p^{\ell}{\wgt}^{2s} \,,
\end{align}
where the inequality follows from Equation \eqref{eq:clique_var22}. Similarly, for a walk $w$ with $\ell = 1$, and $s= k$, we have,
\begin{eqnarray}
{\rm Var} \Big(  \theta (w,G_{\Omega}) {\mathbb I}(w\subseteq G_\Omega)   \Big)
 \leq  f(k) {\wgt}^{2k-1} + g(k)\wgt p^{2-2k}  +p{\wgt}^{2k} \,,  \nonumber
\end{eqnarray}
where we used the inequality in \eqref{eq:clique_var_selfloops}.

For covariance term, we have,
\begin{eqnarray}
	&& {\rm Cov} \Big(  \theta(w_1,G_{\Omega})  \mathbb{I}(w_1 \subseteq G_\Omega)\,,   \theta(w_2,G_{\Omega}) \mathbb{I}(w_2 \subseteq G_\Omega)  \Big) \nonumber \\
	&\leq & \E\Big[ \theta(w_1,G_{\Omega})  \theta(w_2,G_{\Omega}) \Big|  \mathbb{I}(w_1 \subseteq G_\Omega)  \mathbb{I}(w_2 \subseteq G_\Omega) \Big] p^{(\ell_1 + \ell_2 - \tilde\ell)} \nonumber\\
	& \leq & f(\ell',s')p^{((\ell_1+\ell_2 - \tilde\ell)-(s_1+s_2))}\Big((\wgt p)^{(s_1+s_2)} + \wgt p \Big)\,,
\end{eqnarray}
where the inequality follows from Equation \eqref{eq:clique_cov}.


\subsection{Proof of Lemma \ref{lem:clique_var}}
\label{sec:proof_clique_var}
When $G$ is  a union of disjoint cliques, the estimator $\theta(w,G_\Omega)$ defined in \eqref{eq:estimate_theta}
has a compact representation.
This follows from the fact that for any two nodes $i$ and $j$ that are connected in $G_\Omega$,
the neighborhoods of $i$ and $j$ in $G_\Omega$ exactly coincide.
If this happens, then the estimator $\theta(w,G_\Omega)$ simplifies as follows.
Consider a walk $w$ with $s$ self-loops, $k$ edges (including self loops), and $\ell$ distinct nodes.
Define a random integer $\widetilde\tau$ as the degree of a node in the clique that $w$ belongs to in the sampled graph $G_\Omega$,
conditioned on the fact that all nodes in $w$ are sampled (if a walk $w$ is sampled then it must belong to the same clique.).
The randomness comes from the sampling of $\Omega$.
It is straightforward that $\widetilde\tau \sim {\rm Binom}(\wgt-\ell,p)+(\ell-1)$ as
there are already $(\ell -1)$ neighbors from the walk $w$ and
the rest of $((\wgt-1)-(\ell-1))$ nodes are sampled in $\Omega$ with probability $p$, where $\wgt$ is the size of the clique in the original graph $G$. For notational simplification, define a random integer $\tau \equiv \widetilde\tau - (\ell-1)$ that is distributed as $\tau \sim {\rm Binom}(\wgt-\ell,p)$.
In the example in Figure \ref{fig:clique}, for the walk $w=(1,2,2,2,3,3,1)$, we have $\ell = 3$, $s=3$,
$\wgt=7$ and a random instance of  $\widetilde\tau=4$ that is $\tau = 2$.
We claim that the estimator $\theta(w,G_\Omega)$ given in \eqref{eq:estimate_theta} is a function of only $\tau$, $\ell$, $s$ and $p$, and can be simplified as follows, and give
a proof in Section \ref{sec:proof_clique_theta}.
\begin{lemma}
	\label{lem:clique_theta}
	When the underlying $G$ is a union of disjoint cliques and $G_\Omega$ is a subgraph obtained via
	vertex sampling with a probability $p$,
	for a length $k$ walk $w$ in $G_\Omega$ with $\ell$ distinct nodes and $s$ self-loops, we have
	\begin{eqnarray} \label{eq:clique_theta1}
		\theta(w,G_\Omega) &=&   \langle  A^{-1}_{s+1} , \overline{\tau} \rangle \;,
	\end{eqnarray}
	where $\tau \equiv \widetilde\tau - (\ell-1)$ and $\widetilde\tau$ is the degree of any node in the clique that $w$ belongs to in the sampled graph $G_\Omega$,
	$\overline{\tau}=[1,\tau,\tau^2,\ldots,\tau^s]$ is a column vector of monomials of $\tau$ up to degree $s$,
	and $A^{-1}_{s+1}$ is the $(s+1)$-th row of the inverse of the matrix $A\in\reals^{(s+1) \times (s+1)}$ satisfying
$
		A \,\overline{\wgt} \;\; = \;\; \E[\overline{\tau}]
$,
	for $\overline\wgt=[1,(\wgt-1), (\wgt-1)^2, \ldots,(\wgt-1)^s]$, a column vector of monomials of $(\wgt-1)$.
	Further, $A$ is a lower-triangular matrix that depends only on
	$\ell$ and $p$, such that
	$\max_{i\in[s+1]} | (A^{-1})_{s+1,i}| = O(p^{-s})$.
\end{lemma}
In the running example, $s=3$, $\widetilde\tau=4$, and $\ell = 3$, and therefore we have
\begin{eqnarray}
A=\begin{bmatrix}
	1& 0 &0 &0 \\
	p -\ell p & p & 0 & 0 \\
	\ell^2 p^2 - \ell p^2 -\ell p + p & -2\ell p^2 + p^2 +p & p^2 & 0\\
	A_{41}
	& A_{42} & -3\ell p^3 + 3p^3 & p^3
\end{bmatrix}\;,
	\label{eq:defA}
\end{eqnarray}
where $A_{41}=-\ell^3 p^3 + 3\ell^2 p^2 + \ell p^3 -3\ell p^2 -\ell p+p $
and $A_{42}=3\ell^2 p^3 - 6\ell p^2 - p^3 + 3p^2 +p$.
For any $s$, corresponding $A$ can be computed immediately from the moments of a Binomial distribution up to degree $s$.
Since $\overline\wgt = \E[A^{-1}\overline\tau] $, this representation  immediately reveals that
\textcolor{black}{$\E[\theta(w,G_\Omega) | w \subseteq G_{\Omega}] = \E[\langle A_{s+1}^{-1},\overline\tau \rangle]=  (\wgt-1)^s$}. Note that $\tau$ is conditioned on the event that all the nodes in $w$ are sampled.

With this definition, the variance of $\theta(w,G_\Omega)$ can be upper bounded as follows.
We will let $f(\ell,s)$ denote a function over $\ell$ and $s$ that captures the dependence in $\ell$ and $s$ that may change from line to line,
and only track the dependence in $\wgt$ and $p$.
\begin{align}
		  &{\rm Var}\Big( \theta(w,G_{\Omega})\Big| w\subseteq G_\Omega   \Big)
		  \;\; \leq\;\;  f(\ell,s)  \max_{i \in [s+1]} {\rm Var}\big(   (A^{-1})_{s+1,i} \, \tau^{i-1}  \big)  \nonumber\\
		  &\hspace{1cm}  \leq \; f(\ell,s)\, p^{-2s}\, {\rm Var} ( \tau^{s} ) \nonumber\\
		&\hspace{1cm} =  \; f(\ell,s) p^{- 2s}\Big(\E[\tau^{2s}]  - \E[\tau^{s}]^2\Big)  \nonumber\\
		&\hspace{1cm} \leq \; f(\ell,s) p^{ - 2s}\Big((\wgt-\ell)^{2s}p^{2s} +f(s)(\wgt-\ell)^{2s-1}p^{2s-1} \nonumber\\
		&\hspace{1cm}\;\;\;\; +\;\; (\wgt-\ell)p - (\wgt-\ell)^{2s}p^{2s}\Big) \nonumber\\
		 &\hspace{1cm} \leq \; f(\ell,s) \Big( p^{-1}(\wgt-\ell)^{2s-1} + (\wgt-\ell)p^{1-2s} \Big) \;,
\end{align}
where
the first inequality follows from the fact that $\overline\tau$ is an $s+1$ dimensional vector,
the second inequality follows from that fact that
$\max_i | (A^{-1})_{s+1,i}| = O(p^{-s})$ from Lemma \ref{lem:clique_theta}
and $\max_i = {\rm Var}(\tau^{i-1})={\rm Var}(\tau^s)$,
and in the third inequality
we used the fact that $\tau \sim {\rm Binom}(\wgt - \ell,p)$ and a result from \cite{berend2010improved} that $\E[({\rm Binom}(d,p))^s] \leq \sum_{j=1}^s S(s,j) (dp)^j $ where $S(s,j)$ is the Sterling number of second kind. $S(s,s) = 1$, $S(s,1) = 1$ and $S(s,j) \leq f(s)$, for $2 \leq j \leq s-1 $. We also used Jensen's inequality $\E[({\rm Binom}(d,p))^s] \geq \E[({\rm Binom}(d,p))]^s$.
This  proves the desired bound in \eqref{eq:clique_var}.

To prove the bound in \eqref{eq:clique_var_selfloops}, observe that when $\ell=1$, $\tau \sim {\rm Binom}(\wgt-1,p)$, and we can tighten the above set of inequalities
\begin{align}
		  &{\rm Var}\Big( \theta(w,G_{\Omega})\Big| w\subseteq G_\Omega   \Big)
		  \;\; \leq\;\;  g(k) \max_{i \in [k+1]} {\rm Var}\big(   (A^{-1})_{k+1,i} \, \tau^{i-1}  \big)  \nonumber\\
		  &\hspace{1cm}  \leq \; g(k)\, p^{-2k}\, {\rm Var} ( \tau^{k} ) \nonumber\\
		&\hspace{1cm} =  \; g(k) p^{- 2k}\Big(\E[\tau^{2k}]  - \E[\tau^{k}]^2\Big)  \nonumber\\
		&\hspace{1cm} \leq \; g(k) p^{ - 2k}\Big((\wgt-1)^{2k}p^{2k} +f(k)(\wgt-1)^{2k-1}p^{2k-1} \nonumber\\
		& \hspace{1cm} \;\;\;\;\;\;+\;\; (\wgt-1)p - (\wgt-1)^{2k}p^{2k}\Big) \nonumber\\
		 &\hspace{1cm} \leq \; f(k) p^{-1}(\wgt-1)^{2k-1} + g(k) (\wgt-1)p^{1-2k}  \;.
\end{align}

For the covariance term, conditioned on the event that both the walks $w_1$ and $w_2$ are observed, distribution of the random degree integer of each walk is $\tilde\tau_1 = \tilde\tau_2 = \tilde\tau_{12}$, where $\tilde\tau_{12} \sim {\rm Binom}(\wgt -(\ell_1+\ell_2 - \tilde\ell),p) + (\ell_1+\ell_2 - \tilde\ell - 1)$. Therefore, the shifted Binomial random variable of each walk $\tau_1 = \tilde\tau_1 - (\ell_1-1) = \tilde\tau_{12} - (\ell_1-1)$, and $\tau_2 = \tilde\tau_2 - (\ell_2-1) = \tilde\tau_{12} - (\ell_2-1)$
For the walk $w_1$, with number of nodes $\ell_1$, self loops $s_1$, lets denote the matrix $A$ in \eqref{eq:clique_theta1} by $A_1$, and similarly for walk $w_2$, denote it by $A_2$. Then we have,
\begin{align}
& \E\Big[ \theta(w_1,G_{\Omega})  \theta(w_2,G_{\Omega}) \Big|  \mathbb{I}(w_1 \subseteq G_\Omega)  \mathbb{I}(w_2 \subseteq G_\Omega) \Big] \nonumber\\
		  &\leq  f(\ell',s')  \max_{i_1 \in [s_1+1], i_2 \in [s_2 +1]} \E\Big[   (A_1^{-1})_{s_1+1,i_1} \, \tau_{1}^{i_1-1} (A_2^{-1})_{s_2+1,i_2} \, \tau_{2}^{i_2-1}  \Big]\, \label{eq:clique_theta2}\\
		  & \leq  f(\ell',s')p^{-(s_1+s_2)} \E\big[\tau_{1}^{s_1} \tau_{1}^{s_2} \big]\, \label{eq:clique_theta3}\\
		  & \leq  f(\ell',s')p^{-(s_1+s_2)}\Big((\wgt p)^{s_1+s_2} + \wgt p \Big)\,,
\end{align}
where the first inequality follows from the fact that $\overline\tau_1$ is an $s_1+1$ dimensional vector, and  $\overline\tau_2$ is an $s_2+1$ dimensional vector,
the second inequality follows from that fact that
$\max_i | (A_1^{-1})_{s_1+1,i_1}| = O(p^{-s_1})$ from Lemma \ref{lem:clique_theta}
and $\max_{i_1,i_2} = \E[\tau_1^{i_1-1} \tau_2^{i_2-1}]=\E[\tau_1^{s_1} \tau_2^{s_2}]$,
and in the third inequality
we used a result from \cite{berend2010improved} that $\E[({\rm Binom}(d,p))^s] \leq \sum_{j=1}^s S(s,j) (dp)^j $ where $S(s,j)$ is the Sterling number of second kind. $S(s,j) \leq f(s)$, for $1 \leq j \leq s$. This  proves the desired bound in \eqref{eq:clique_cov}.


\subsection{Proof of Lemma \ref{lem:clique_theta}}
\label{sec:proof_clique_theta}
We are left to prove that
the estimator  $\theta(w,G_\Omega)$ simplifies as in \eqref{eq:clique_theta1}
when the original graph is a clique graph(or a union of disjoint cliques).

We use the notations introduced in Section  \ref{sec:schatten_theta} and Section \ref{sec:proof_clique_var}.  Consider a closed walk $w$ of length $k$ on $\ell$ distinct nodes with $U = \{u_1,\cdots,u_{\tilde\ell}\}$ set of nodes in it that have at least one self-loop, $|U| = \tilde\ell$, and a total of $s$ self loops. If the underlying graph is a clique graph the partition of $V$ defined in \eqref{eq:partitions}, for any $T \subseteq U$, is as follows:
\begin{eqnarray}\label{eq:same1}
V_{T,U\setminus T} =
\begin{cases}
\emptyset &\text{if }  |T| < |U| -1\\
d_{U,\emptyset} & \text{if }  T = U\\
v & \text{if }  T = U\setminus v, \text{ for any } v \in U\,.
\end{cases}
\end{eqnarray}
Recall that $d_{U,\emptyset} \equiv \cap_{v \in U}\partial v$.
Therefore, we have,
\begin{eqnarray}\label{eq:same2}
d_{T,U\setminus T}(w) =
\begin{cases}
0 &\text{if }  |T| < |U| -1\\
\ell-\tilde\ell & \text{if }  T = U\\
1 & \text{if } T = U\setminus v, \text{ for any } v \in U\,.
\end{cases}
\end{eqnarray}
If the underlying clique graph is of size $\wgt$ then $|d_{U,\emptyset}| = \wgt - \tilde\ell$. Using the fact  $d_{T,U\setminus T}(\Omega) \sim {\rm Binom}(d_{T,U\setminus T}-d_{T,U\setminus T}(w),p) + d_{T,U\setminus T}(w)$, as explained in Section \ref{sec:schatten_theta}, we have,
\begin{eqnarray*}\label{eq:same3}
d_{T,U\setminus T}(\Omega) \sim
\begin{cases}
0 &\text{if }  |T| < |U| -1\\
 {\rm Binom}(\wgt-\ell,p) + (\ell-\tilde\ell) & \text{if }  T = U\\
1 & \text{if } T = U\setminus v, \text{ for any } v \in U\,.
\end{cases}
\end{eqnarray*}
Using Equation \eqref{eq:same1} it is immediate that that degree of any node $u \in U$ is $d_u = d_{U,\emptyset} + (\tilde\ell -1)$, and hence,
$\E[\theta(w,G_{\Omega})| w \in G_{\Omega}] = \big(d_{U,\emptyset} + (\tilde\ell -1)\big)^s = (\wgt-1)^s$. Therefore, an alternative characterization of the estimator defined in \eqref {eq:estimate_theta} is as following: $\theta(w,G_{\Omega})$, conditioned on the event that all the nodes in the walk $w$ are sampled, is a random variable dependent only upon $d_{U,\emptyset}(\Omega) \sim  {\rm Binom}(\wgt-\ell,p) + (\ell-\tilde\ell)$ such that its conditional expectation is $\E[\theta(w,G_{\Omega})| w \in G_{\Omega}] = (\wgt-1)^s$. With change of notations it is immediate that the estimator defined in \eqref{eq:estimate_theta} is same as the estimator in \eqref{eq:clique_theta1} when the underlying graph is a clique graph(or a disjoint union of cliques).

By the definition of $A$, it follows that
the diagonal entries are exactly $\diag([1,p,\ldots,p^s])$ and
the bottom-left off-diagonal entris are all $\Theta(1)$ with respect to $p$,
and the top-right off-diagonal entries are all zeros.
Applying the inverse to this lower triangular matrix, it follows that
$A^{-1}$ is also a lower triangular matrix with diagonal entries
$\diag([1,p^{-1},\ldots,p^{-s} ])$ and the bottom-left off-diagonal entries  are
all $\Theta(1)$. It follows that  $\max_{i\in[s+1]} | (A^{-1})_{s+1,i}| = O(p^{-s})$.

\subsection{Proof of Theorem \ref{thm:main_ER}}
\label{sec:main_proof}
The following lemma provides bound on variance of  Schatten $k$-norm estimator  for a connected general graph with maximum degree $\dmax$. We provide a proof in Section \ref{sec:general_graph1}
\begin{lemma} For a connected graph $G$ on $\wgt$ vertices with maximum degree $\dmax$, variance of Schatten $k$-norm estimator $\hTheta_k(G_\Omega)$ is bounded by
\begin{eqnarray} \label{eq:lem_ER1}
{\rm Var} \big(\hTheta_k(G_\Omega)\big) & \leq & h(k) \frac{\wgt^2 \dmax^{2k}}{p}\Big( 1 + \frac{1}{(\dmax p)^{2k-1}}\Big)\,,
\end{eqnarray}
where $h(k) = O(2^{k^2})$. Moreover, if there exists a positive constant $c$ such that $\dmax^3 p \geq c 2^{k^2}$ or $\dmax p^3 \leq c/2^{k^2}$ then \eqref{eq:lem_ER1} holds with $h(k) = {\rm poly}(k)$. \label{lem:ER}
\end{lemma}
Using Equations \eqref{eq:mse} and \eqref{eq:break_var} along with Lemma \ref{lem:ER}, Theorem \ref{thm:main_ER} follows immediately.

\subsection{Proof of Lemma \ref{lem:ER}}
\label{sec:general_graph1}

We use the notations introduced in Section \ref{sec:schatten_theta} and Section \ref{sec:clique}. Denote the size of the connected component by $\wgt$ and let $\dmax$ be the maximum degree of any node in the connected component.

The following technical lemma provides upper bounds on the variance and covariance terms.
We provide a proof in Section \ref{sec:ER_var_0}.

\begin{lemma}
	\label{lem:ER_var_0}
Under the hypothesis of Lemma \ref{lem:ER}, for a length-$k$ walk $w$ over $\ell$ distinct nodes with $s \geq 1$ self-loops, the following holds:
	\begin{eqnarray}
{\rm Var} \Big(  \theta (w,G_{\Omega}) {\mathbb I}(w\subseteq G_\Omega)   \Big)
 \leq  h(k) \Big( p^{\ell}{\dmax}^{2s} + \dmax p^{\ell+1-2s} \Big)   ,
 \label{eq:ER_var}
\end{eqnarray}
and when $\ell =1$, $s = k$, we have,
	\begin{align}
& \;{\rm Var} \Big(  \theta (w,G_{\Omega}) {\mathbb I}(w\subseteq G_\Omega)   \Big) \nonumber\\
& \;\;\;\; \leq  f(k) {\dmax}^{2k-1} \;+\; g(k)\dmax p^{2-2k}  \;+\; p{\dmax}^{2k} \,, \label{eq:ERpf8}
\end{align}
and for any length-$k$ walks $w_1, w_2$ over $\ell_1, \ell_2$ distinct nodes  with $\tilde\ell$ unique overlapping nodes, $|w_1 \cap w_2| = \tilde\ell $, $s_1,s_2 \geq 1$ self-loops respectively, the covariance term can be upper bounded by:
\begin{eqnarray}
	{\rm Cov} \Big(  \theta(w_1,G_{\Omega})  \mathbb{I}(w_1 \subseteq G_\Omega)\,,   \theta(w_2,G_{\Omega}) \mathbb{I}(w_2 \subseteq G_\Omega)  \Big)  \nonumber \\
	\leq  h(k)p^{((\ell_1+\ell_2 - \tilde\ell)-(s_1+s_2))}\Big((\dmax p)^{(s_1+s_2)} + \dmax p \Big)\,, \label{eq:ERpf7}
\end{eqnarray}
for some function $h(k) = O(2^{k^2})$, $f(k)=O(k!)$, and $g(k) = {\rm poly}(k)$, where $p$ is the vertex sampling probabiliy.
\end{lemma}

The total count of length $k$ closed cycles on $\ell$ distinct nodes in a general graph on $\wgt$ nodes graph with maximum degree $\dmax$ is bounded by $f(k)\wgt \dmax^{\ell-1}$. It follows from the observation that fixing a node in the cycle, there are at most $\dmax^{\ell-1}$ paths to $\ell-1$-hop neighbors. That is $|w \in W : \cH(w) \in  \bH_{k,\ell,s}| \leq f(k)\wgt \dmax^{\ell-1}$ for any $1 \leq s \leq k$.

We use these inequalities to get bound on variance and covariance terms in \eqref{eq:pf2}.

For a walk $w \in W$ with $\cH(w)\in \bH_{k,\ell,s}$ with $1 \leq s \leq k-2$, using \eqref{eq:ER_var}, we have,
\begin{align}
& \frac{\wgt \dmax^{\ell-1}}{p^{2\ell}}{\rm Var} \Big(  \theta(w,G_{\Omega}) {\mathbb I}(w\subseteq G_\Omega)\Big) \nonumber\\
& \;\;\;\; \leq  h(k) \Big( \frac{{\dmax}^{\ell+2s-1}}{p^{\ell}} + \frac{\dmax^{\ell}}{p^{\ell + 2s -1}} \Big)  \nonumber\\
& \;\;\;\;\leq  h(k) \wgt \Big(\frac{\dmax^{2k-3}}{p^2} + \frac{\dmax^{2}}{p^{2k -3}} \Big) \,. \label{eq:ERpf9}
\end{align}

For a walk $w \in W$ with $\cH(w)\in \bH_{k,\ell,s}$ with $\ell=1$, $s = k$, using \eqref{eq:ERpf8}, we have,
\begin{align}
&\frac{\wgt}{p^{2\ell}}{\rm Var}\Big(  \theta(w,G_{\Omega}) {\mathbb I}(w\subseteq G_\Omega)\Big) \;\leq \nonumber\\
& \;\;\;\; f(k) \wgt{\dmax}^{2k-1}p^{-2} \;+\; g(k)\wgt\dmax p^{-2k} \;+ \; \wgt\dmax^{2k}p^{-1} \,. \label{eq:ERpf10}
\end{align}

For a walk $w$ with $s=0$, $ \theta(w,G_{\Omega}) = 1$, and, we have,
\begin{eqnarray}
\frac{\wgt\dmax^{\ell-1}}{p^{2\ell}}{\rm Var}\Big(  \theta(w,G_{\Omega}) {\mathbb I}(w\subseteq G_\Omega)\Big) & \leq & \frac{\wgt\dmax^{\ell-1}}{p^{\ell}}\,. \label{eq:ERpf11}
\end{eqnarray}

Combining, Equations \eqref{eq:ERpf9}, \eqref{eq:ERpf10}, and \eqref{eq:ERpf11}, and using $|w \in W : \cH(w) \in  \bH_{k,\ell,s}| \leq f(k)\wgt \dmax^{\ell-1}$, we have
\begin{align}
&\sum_{\ell = 1}^k  \sum_{s = 0}^{k-\ell+1} \sum_{H \in \bH_{k,\ell,s}} \Big\{  \frac{1}{p^{2\ell}} \, \sum_{w \in W: \cH(w)=H} \mathbb{I}(w \subseteq G)\, \times \nonumber \\
& \;\;\;\; \;\; {\rm Var} \Big(  \theta(w,G_{\Omega}) {\mathbb I}(w\subseteq G_\Omega)   \Big) \Big\} \nonumber\\
& \leq  h(k)\wgt\dmax p^{-2k} \;+\; \wgt\dmax^{2k} p^{-1}\,,\label{eq:pf105}
\end{align}
and if $\dmax p^3 \leq 1/h(k)$, then the above quantity is bounded by $g(k)\wgt \dmax p^{-2k}(1 + o(1))$.
If $\dmax^3 p \geq h(k)$, then the above quantity is bounded by $\wgt\dmax^{2k}p^{-1}(1 + o(1))$.

Analysis of covariance terms in \eqref{eq:pf2} follows along the similar lines as that of the clique graph case and the result in Lemma \ref{lem:ER} follows immediately.

\subsection{Proof of Lemma \ref{lem:ER_var_0}}
\label{sec:ER_var_0}
We give a lemma similar to Lemma \ref{lem:clique_var} for the case of a general graph that provides a bound on conditional variance and conditional covariance terms. We give a proof in Section \ref{sec:cond_var_ER}.

\begin{lemma}
	\label{lem:cond_var_ER}
	Under the hypothesis of Lemma \ref{lem:ER}, for length-$k$ walks $w_1, w_2$ over $\ell_1, \ell_2$ distinct nodes with $s_1,s_2 \geq 1$ self-loops respectively,
	the conditional variance of estimator $\theta(w_1,G_\Omega)$, defined in  \eqref{eq:estimate_theta}, given that all the nodes in the walk are sampled
	can be upper bounded by
	\begin{align}
		 &{\rm Var}\Big( \theta(w_1,G_{\Omega})\Big| w_1\subseteq G_\Omega   \Big)
		 \;\;\leq \;\; \nonumber\\
		  & \;\;\;\;\;\;\; h(k)\Big(\dmax^{2s_1} + \dmax p^{1-2s_1}\Big) \;, \text{ and } \label{eq:ER_var22}\\
		  &\E\Big[ \theta(w_1,G_{\Omega})  \theta(w_2,G_{\Omega}) \Big|  \mathbb{I}(w_1 \subseteq G_\Omega)  \mathbb{I}(w_2 \subseteq G_\Omega) \Big]
		  \;\;\leq \nonumber\\
		  & \;\;\;\;\;\;\; 	h(k) p^{-(s_1+s_2)}\Big((\dmax p)^{s_1+s_2} + \dmax p \Big) \;,
 	\label{eq:ER_cov}
 	\end{align}
		for some function $h(k)=O(2^{k^2})$,
		where $p$ is the vertex sampling probability. Moreover, for a length $k$ walk $w$ with $\ell = 1$, and $s= k$,
			\begin{eqnarray}
		 {\rm Var}\Big( \theta(w,G_{\Omega})\Big| w\subseteq G_\Omega   \Big)
		 		  \leq   f(k)  p^{-1}\dmax^{2k-1} + g(k) \dmax p^{1-2k} \,,  \label{eq:ER_var_selfloops}
    \end{eqnarray}
    for some function $f(k) = O(k!)$, and $g(k) = {\rm poly}(k)$.
\end{lemma}
Using the above lemma, proof of Lemma \ref{lem:ER_var_0} follows along the lines of the  proof of Lemma \ref{lem:clique_var_0}.

\subsection{Proof of Lemma \ref{lem:cond_var_ER}}
\label{sec:cond_var_ER}
Recall that for a general graph $\theta(w,G_{\Omega})$ is an unbiased estimator of $\prod_{u\in w} d_u^{s_u}$ and is given in \eqref{eq:estimate_theta}. It is easy to see that for any given walk $w$ on $\ell$ distinct nodes and with $s$ self-loops,
\begin{eqnarray}
{\rm Var}\Big(\theta(w,G_{\Omega}) | w \subseteq G_{\Omega}\Big)  \leq   h(k) \max_{\bT}\Bigg\{{\rm Var} \bigg(\Big\{
		\prod_{T\in \bT}      \widehat{d}_{T,U\setminus T}^{(t_T)}
		\Big\}\bigg)\Bigg\}\,,		\nonumber
\end{eqnarray}
where $h(k) = O(2^{k^2})$. It follows from the fact that there are at most ${k/2}$ distinct nodes with self loops and hence at most $2^{k/2 -1}$ partitions in \eqref{eq:partitions} which leads to at most $2^{k^2/4}$ summation terms in \eqref{eq:sumdegree2}. Further $\prod_{T\in \bT}      \widehat{d}_{T,U\setminus T}^{(t_T)}$ is the product of independent random variables. Observe that using Lemma \ref{lem:clique_var}, we have
\begin{eqnarray}
{\rm Var}\Big( \widehat{d}_{T,U\setminus T}^{(t_T)}\Big) & \leq & f(k)\Big(p^{-1}\dmax^{2t_T-1} + \dmax p^{1-2t_T} \Big)\,
\end{eqnarray}
and $\E[\widehat{d}_{T,U\setminus T}^{(t_T)}] \leq \dmax^{t_T}$.
Using the fact that for independent random variables $X_1,X_2,\cdots,X_n$,
\begin{eqnarray}
{\rm Var}(X_1 X_2\cdots X_n) = \prod_{i=1}^n \Big({\rm Var}(X_i) + (\E[X_i])^2\Big) - \prod_{i=1}^n (\E[X_i])^2\,,
\end{eqnarray}
we have,
\begin{align}
	& \max_{\bT}\Bigg\{{\rm Var} \bigg(\Big\{
		\prod_{T\in \bT}      \widehat{d}_{T,U\setminus T}^{(t_T)}
		\Big\}\bigg)\Bigg\}  \nonumber\\
		&\;\;\;\;\;\; \leq
		\prod_{T \in \bT} f(k)\Big(p^{-1}\dmax^{2t_T-1} + \dmax p^{1-2t_T}  + \dmax^{2t_T}\Big) \nonumber \\
		& \;\;\;\;\;\; \leq  f(k)\Big(\dmax^{2s} + \dmax p^{1-2s}\Big)\,
\end{align}
where in the last inequality we used that $\sum_{T \in \bT} t_T = s$.  \eqref{eq:ER_var22} follows from collecting the above inequalities.   \eqref{eq:ER_cov} follows from the definition of $\theta(w,G_{\Omega})$ given in \eqref{eq:estimate_theta} and the proof of  \eqref{eq:clique_cov} of Lemma \ref{lem:clique_var}.
 \eqref{eq:ER_var_selfloops} follows directly from \eqref{eq:clique_var_selfloops} of Lemma \ref{lem:clique_var}.


\subsection{Proof of Proposition \ref{pro:poly}}
\label{sec:poly_proof}

\begin{align}
 	&\max_{x\in[\alpha,1]} |H_\alpha(x) - f_{\bb^*}(x)|  \leq  \max_{x\in[\alpha,1]} |H_\alpha(x) - f_{\tb}(x)|\,, \\
 	&\;\;\;\; =  \max\Big\{|H_\alpha(\alpha) - f_{\tb}(\alpha)|, |H_\alpha(1) - f_{\tb}(1)|\Big\}\\
 	&\;\;\;\; =  \bigg(\frac{1-\alpha}{1+\alpha}\bigg)^m\,
	\label{eq:poly}
\end{align}
where $\tb\equiv (2/(1+\alpha))[1,\ldots,1]$.

\end{document}